\documentclass{article}

\PassOptionsToPackage{numbers, compress}{natbib}

\usepackage[final,main]{neurips_2026}

\usepackage[utf8]{inputenc} 
\usepackage[T1]{fontenc}    
\usepackage[hidelinks]{hyperref}       
\usepackage{url}            
\usepackage{booktabs}       
\usepackage{amsfonts}       
\usepackage{nicefrac}       
\usepackage{microtype}      
\usepackage{xcolor}         

\usepackage{siunitx}
\usepackage{wrapfig}
\usepackage{multirow}
\usepackage{adjustbox}
\usepackage{enumitem}
\usepackage{pifont}
\usepackage{algorithm}
\usepackage{algpseudocode}
\usepackage{graphicx}
\usepackage{amsmath}
\usepackage{amssymb}

\title{FRUC: Feedforward Dynamic Scene Reconstruction from Uncalibrated Collaborative Driving Views}

\author{
  Yihang Tao,
  Yu Guo,
  Zhengru Fang\thanks{Corresponding author.}, 
  Haonan An,
  Yuguang Fang \\
  Hong Kong JC STEM Lab of Smart City, City University of Hong Kong \\
  \texttt{\{yihang.tommy, yu.guo, zhefang4-c, haonanan2-c\}@my.cityu.edu.hk} \\
  \texttt{my.Fang@cityu.edu.hk}
}

\begin{document}

\maketitle

\begin{abstract}
We present \texttt{FRUC}, a feedforward 3D Gaussian Splatting framework for dynamic scene reconstruction from uncalibrated collaborative driving views. 
Existing multi-agent reconstruction frameworks are often hindered by rigid prerequisites, demanding precise spatial calibration and slow per-scene optimization. 
In this paper, we rethink this task by conceptualizing a distributed multi-vehicle network as a spatio-temporally unstructured ego-centric multi-camera system, where the core challenge lies in enhancing ego-centric occluded geometry through collaboration without degrading the ego's accurately observed visible geometry, while preserving reconstruction efficiency.
For efficient reconstruction, \texttt{FRUC} is built upon a visual grounded geometric Transformer backbone to enable one-shot, calibration-free inference from a flexible number of multi-vehicle views. 
To achieve non-destructive geometric supplementation under uncalibrated cross-agent misalignment, \texttt{FRUC} first introduces an ego-centric causal occlusion field that explicitly derives occlusion evolution as latent priors by modeling agent-wise spatio-temporal correlations. 
Guided by these occlusion priors, it further formulates cross-agent integration as a deterministic residual denoising process via zero-initialized injection, turning challenging cross-agent fusion into bounded residual learning for robust collaborative blind-spot completion. Through extensive evaluations on real-world V2X-Real and UrbanIng-V2X datasets, \texttt{FRUC} is shown to be a new state-of-the-art for the scene reconstruction of dynamic collaborative driving environments, significantly outperforming existing methods in both rendering quality and efficiency.
Code is available at \url{https://github.com/yihangtao/FRUC.git}.
\end{abstract}

\section{Introduction}\label{sec:introduction}

Fast and scalable dynamic scene reconstruction forms the bedrock of modern autonomous driving, powering both on-board execution and cloud-based simulation. On-board, rapidly lifting sparse historical observations into 4D representations enables real-time what-if analysis. This allows systems to synthesize novel views and simulate potential future states for critical decision-making \citep{tian2025drivingforward, hou2025drivingscenemultitaskonlinefeedforward, chen2025dggtfeedforward4dreconstruction, Wu_2024_CVPR, jiang2025anysplat, Fan_2025_CVPR}. In the cloud, efficiently transforming extensive driving logs into large-scale, dynamic digital twins provides the robust closed-loop simulators essential for training generalizable driving world models \citep{zhou2024drivinggaussian, xiong2023sparsegs, kerbl20233d, lu2024drivingreconlarge4dgaussian, luiten2023dynamic, shao2023tensor4d}. 
To meet the efficiency demanded by these continuous sensory streams, feedforward 3D Gaussian Splatting (3DGS) has emerged as the premier single-pass solution. Consequently, recent single-vehicle pipelines have rapidly evolved from efficient surround-view static reconstruction \citep{tian2025drivingforward} and spatial-temporal dynamic modeling \citep{yang2024stormspatiotemporalreconstructionmodel} to unified frameworks supporting pose-free dynamic scene reconstruction directly from unposed images \citep{zuo2025dvgt, chen2025dggtfeedforward4dreconstruction}.

\begin{figure*}[t]
  \centering
    \includegraphics[width=\textwidth, keepaspectratio]{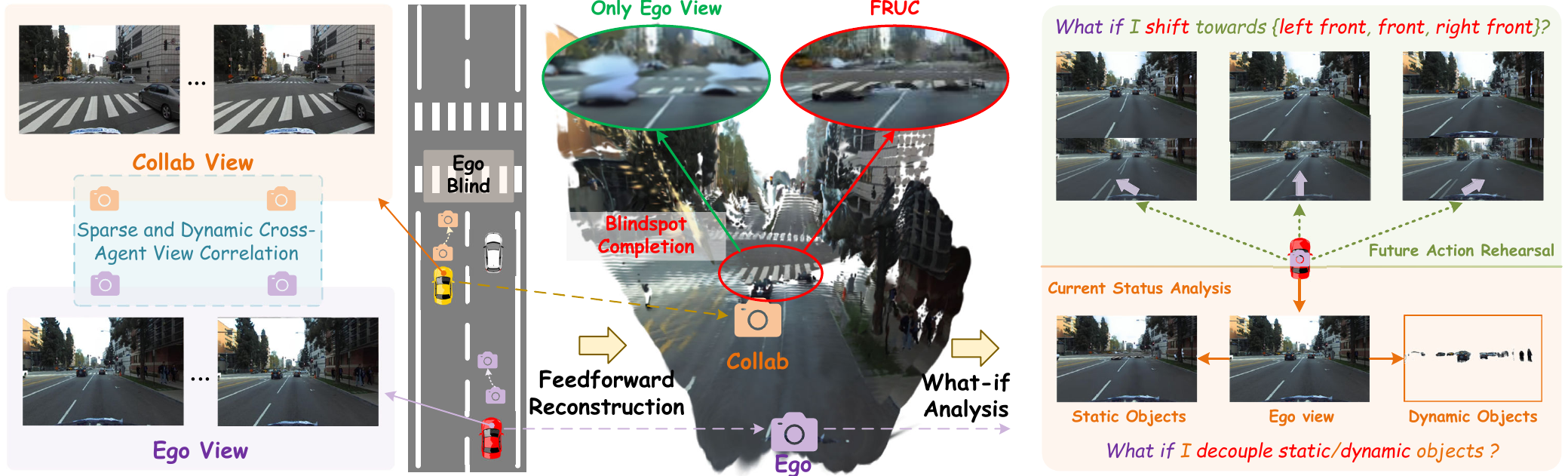}
  \vspace{-6mm}
  \caption{\textbf{Collaborative driving scene reconstruction and what-if analysis with \texttt{FRUC}.} By aggregating uncalibrated, sparse, and dynamic cross-agent observations, \texttt{FRUC} seamlessly completes the ego vehicle's blind spots via a single feedforward pass. It provides accurate occluded background recovery, leading to reliable scene editing and high-fidelity novel view synthesis.}
  \label{fig:teaser}
  \vspace{-4mm}
\end{figure*}

Despite these rapid advancements, ego-centric reconstruction remains fundamentally bottlenecked by the physical line-of-sight \citep{xu2025cruisecooperativereconstructionediting, 11097436, NEURIPS2021_f702defb, fangPACPPriorityAwareCollaborative2024, 11207636, huWhere2commCommunicationefficientCollaborative2024, 11127818, tao2026learningmutualviewinformation}. Severe visual occlusions in single-vehicle perspectives inevitably result in incomplete scene geometry, which critically constrains what-if analysis. For instance, when removing a foreground object to synthesize novel safety-critical scenarios for closed-loop evaluation, single-vehicle representations invariably reveal unobserved ``holes'' in the background, severely degrading simulation fidelity. Collaborative driving systems offer a paradigm shift to shatter this physical limitation. By aggregating complementary viewpoints from distributed agents, collaborative reconstruction can effectively ``see through'' occlusions to establish a holistic and geometrically complete 3D environment (as illustrated in Fig.~\ref{fig:teaser}). Recently, methods like CRUISE \citep{xu2025cruisecooperativereconstructionediting} and V2X-Gaussians \citep{11097436} pioneered this domain. However, these collaborative methods rely heavily on rigid prerequisites, assuming that multi-source agent cameras are well-calibrated and often requiring auxiliary sensor data (e.g., LiDAR point clouds) for spatial initialization. Furthermore, they depend on computationally expensive per-scene optimization. This rigid dependency makes them notoriously slow and fundamentally impractical for real-time, scalable applications.

To overcome these limitations, we re-formulate this task as \textit{feedforward reconstruction from uncalibrated collaborative driving views}. However, extending ego-centric models to multi-agent scenarios introduces two formidable challenges due to the inherent instability of collaborative networks: \ding{182} \textit{Unstructured Geometric Misalignment:} Single-vehicle multi-camera systems possess fixed relative extrinsics, allowing straightforward fusion via deterministic warping \citep{tian2025drivingforward}. Conversely, multi-agent systems introduce uncalibrated, time-varying, and dynamically moving cameras. This lack of precise cross-agent calibration renders simple geometric projection largely ineffective for aligning these highly unstructured viewpoints. \ding{183} \textit{Destructive Semantic Interference:} Single-vehicle sequences provide a highly stable spatiotemporal context with constant spatial overlap and smooth temporal transitions between consecutive frames, offering robust cues for 3D geometry inference. Introducing external cameras drastically destabilizes this pattern: spatio-temporal overlap becomes unpredictable, and cross-vehicle semantic correlations drop significantly. Forcing a network to jointly encode this highly irregular context creates a cluttered latent space, which overwhelms the network and severely corrupts the ego vehicle's previously established geometric priors, even destroying its reliable single-vehicle reconstruction capabilities  (e.g., ghosting or distorted structures).

In this paper, we introduce \texttt{FRUC}, a unified feedforward framework designed to tackle these challenges. Rather than employing rigid geometric warping or naive feature concatenation, we conceptually re-formulate cross-agent integration as a bounded residual learning problem. Built upon a visual geometric grounded Transformer (VGGT) backbone \citep{wang2025vggt}, \texttt{FRUC} extracts ego-centric occlusion priors to guide a deterministic latent residual denoising process. This high-level design implicitly handles uncalibrated multi-agent misalignment and robustly supplements dynamic blind spots, while preserving the ego's reliable observations from destructive cross-agent semantic interference.

Our main contributions are summarized as follows:
\begin{itemize}[leftmargin=10pt]
\item We propose \texttt{FRUC}, a unified feedforward 3DGS framework designed to seamlessly handle spatio-temporally unstructured multi-camera inputs for collaborative driving systems, achieving highly efficient one-shot inference without relying on precise multi-agent calibration.
\item To achieve non-destructive geometric supplementation, we introduce an ego-centric causal occlusion field that explicitly captures occlusion evolution, coupled with a cross-agent latent residual denoising process via prior injections. This enables robust assimilation of collaborative blind-spot information while preserving the ego vehicle's reliably observed visible geometry.
\item Extensive evaluations based on the real-world V2X-Real and UrbanIng-V2X datasets demonstrate that \texttt{FRUC} establishes a new state-of-the-art for collaborative driving scene reconstruction, significantly outperforming existing optimization-based and single-agent feedforward methods in both rendering quality and efficiency.
\end{itemize}

\section{Related Work}\label{sec:related_work}

\subsection{Feedforward 3D Reconstruction}
Recent advancements in 3D Gaussian Splatting (3DGS) \citep{kerbl20233d} have sparked significant interest in feedforward reconstruction methods that eliminate the need for tedious per-scene optimization by learning powerful priors from large-scale datasets. Early approaches, such as Splatter Image \citep{szymanowicz24splatter} and GS-LRM \citep{gslrm2024}, demonstrate the feasibility of predicting 3DGS parameters in a single forward pass, but they primarily focus on object-centric or indoor static scenes. 
Subsequent works like pixelSplat \citep{charatan23pixelsplat} and MVSplat \citep{chen2024mvsplat} leverage multi-view stereo concepts to improve geometry prediction. AnySplat \citep{jiang2025anysplat} attempts to lift unconstrained multi-view captures into 3D Gaussians. However, these methods typically rely on dense multi-view inputs with substantial visual overlap to construct cost volumes or establish geometry priors. Consequently, they struggle to generalize to large-scale, outdoor driving environments, where onboard surround-view cameras naturally exhibit minimal overlap.

\subsection{Multi-view Driving Scene Reconstruction}
To address the unique demands of autonomous driving, recent efforts have shifted towards feedforward architectures tailored for large-scale, outdoor scenes. DrivingForward \citep{tian2025drivingforward} proposes a real-time framework by jointly training a pose and a depth network to predict Gaussian primitives from flexible surround-views. STORM \citep{yang2024stormspatiotemporalreconstructionmodel} aggregates 3D Gaussians from sparse observations using self-supervised scene flows, transforming them across multiple timesteps for dynamic reconstruction. More recently, DGGT \citep{chen2025dggtfeedforward4dreconstruction} and StreetForward \citep{yu2026streetforward} leverage grounded transformers and cross-view attention to efficiently reconstruct dynamic driving environments directly from unposed image sequences, bypassing explicit geometric priors. 
Despite their impressive performance, they are limited to the singe-vehicle setting, which is inherently constrained by the ego vehicle's limited physical line-of-sight, and inevitably suffers from severe geometric incompleteness in occluded regions. Besides, these methods are tailored for fixed multi-camera view relationships within a single vehicle, limiting its applicability to unstructured and dynamic multi-vehicle multi-camera systems.

\subsection{Collaborative Driving Scene Reconstruction}
Collaborative driving systems aggregate distributed viewpoints to overcome the inherent physical limitations (e.g., occlusions) of single-vehicle perception. Recently, optimization-based 3DGS frameworks have been introduced to this domain. CRUISE \citep{xu2025cruisecooperativereconstructionediting} proposes a cooperative reconstruction and editing pipeline that relies heavily on dense 3D annotations to separate dynamic vehicles for novel scene synthesis. Meanwhile, V2X-Gaussians \citep{11097436} introduces a V2X-tailored method that computes exact geometric ray intersections to identify overlapping areas. Despite their success, these methods share critical limitations when faced with real-world unconstrained V2X networks: they assume well-calibrated mult-agent observations, heavily rely on multi-modal inputs (e.g., LiDAR) for labeling or initialization, and operate on a computationally prohibitive per-scene optimization paradigm. In contrast, our \texttt{FRUC} is the first feedforward framework that elegantly bypasses these rigid constraints, implicitly learning cross-agent alignment to achieve highly efficient, single-pass dynamic scene reconstruction directly from uncalibrated collaborative driving views.
\section{Methodology}\label{sec:method}

\begin{figure*}[t]
  \centering
  \includegraphics[width=\textwidth, keepaspectratio]{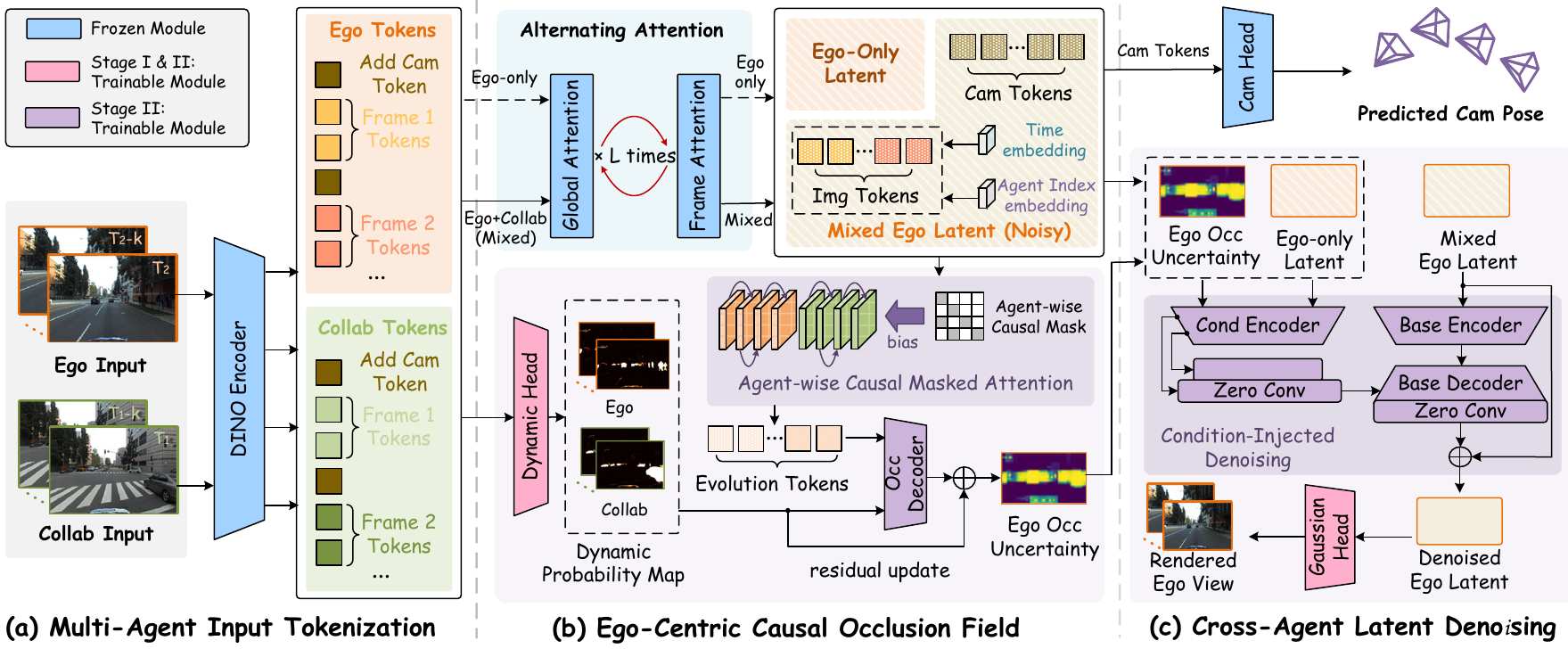} 
  \vspace{-6mm}
  \caption{\textbf{The proposed \texttt{FRUC} framework.} It reconstructs collaborative driving scenes by extracting uncalibrated multi-agent context, inferring dynamic spatial priors through an ego-centric causal occlusion field, and integrating collaborative geometry via cross-agent latent residual denoising.}
  \label{fig:pipeline}
  \vspace{-2mm}
\end{figure*}

We conceptualize a distributed multi-vehicle network as a spatio-temporally unstructured ego-centric multi-camera system, bypassing the need for exact camera extrinsics and hardware-level time synchronization. Given a specific spatial region, we assume a joint set of temporally ordered, uncalibrated multi-agent input images $\{\mathbf{I}_k^\tau \mid \mathbf{I}_k^\tau \in \mathbb{R}^{H \times W \times 3}, \tau = 1, \dots, N\}$ captured by an arbitrary agent $k \in \{e, c\}$ (where $e$ denotes ego and $c$ denotes collab) over $N$ timestamps. Our objective is to reconstruct a temporally coherent, geometrically complete 4D representation in a single forward pass, as conceptually illustrated in Fig.~\ref{fig:pipeline}. 

\subsection{Multi-Agent View Input Tokenization}\label{sec:collab_tokenization}
To process the aforementioned $N$-timestamp multi-agent sequence, we first flatten and tokenize the inputs. For clarity, we use $N=2$ timestamps $\tau \in \{t_0, t_1\}$ for illustration. We concatenate the raw images into a unified multi-agent sequence $\mathbf{I} = [\mathbf{I}_{e}^{t_0}, \mathbf{I}_{c}^{t_0}, \mathbf{I}_{e}^{t_1}, \mathbf{I}_{c}^{t_1}]$. Each frame is then encoded into $L$ high-level patchified tokens $\mathbf{F}_{dino, k}^\tau$ of dimension $D$ using a DINO encoder \citep{oquab2023dinov2}.

To process these collaborative perceptual streams without relying on rigid global coordinate systems, we formulate the distributed observations as a joint set of semantic proxies. Specifically, following the VGGT architecture \citep{chen2025dggtfeedforward4dreconstruction}, a learnable camera token $\mathbf{c}_{cam}$ is appended to the DINO tokens of each frame. These augmented sequences are then processed through the VGGT Alternating-Attention (AA) feature backbone, which aggregates inter-frame and intra-frame contexts to infer the implicit multi-view geometry:
\begin{equation}
\mathbf{F}_{m, k}^\tau, \hat{\mathbf{c}}_k^\tau = \mathcal{E}_{vggt}\Big( [ \mathbf{c}_{cam}, \mathbf{F}_{dino, k}^\tau ] \Big),
\end{equation}
where $\mathbf{F}_{m, k}^\tau \in \mathbb{R}^{L \times D}$ represents the aggregated deep image features absorbing rich multi-view context, and $\hat{\mathbf{c}}_k^\tau$ is the refined implicit camera token encapsulating the source view's spatial prior. 

While the resulting deep features establish a powerful rendering manifold, this coarse fusion inherently creates an entangled latent space where uncalibrated collaborative perspectives interferes the ego's geometric representation. To facilitate motion-aware downstream alignment, we augment the aggregated tokens with explicit temporal and identity information. Specifically, we introduce a learnable agent-identity embedding $\mathbf{E}_{agt}(k) \in \mathbb{R}^D$ and add it to a normally distributed temporal embedding $\mathbf{E}_{tem}(\tau) \in \mathbb{R}^D$, corresponding to the discrete frame timestamp $\tau$:
\begin{equation}\label{eq:meta_embed}
\mathbf{E}_{meta}(k, \tau) = \mathbf{E}_{agt}(k) + \mathbf{E}_{tem}(\tau) \in \mathbb{R}^{D}.
\end{equation}
We then augment every patch token in the frame with this joint meta-data vector via broadcasting, leading to $\tilde{\mathbf{F}}_{m, k}^\tau = \mathbf{F}_{m, k}^\tau + \mathbf{E}_{meta}(k, \tau) \in \mathbb{R}^{L \times D}$. This explicit encoding of cross-agent identity and temporal causality serves as spatio-temporal anchors during the subsequent feature denoising. For simplicity, we denote the full sequence of augmented tokens as $\tilde{\mathbf{F}}_{m}$.

\subsection{Ego-Centric Causal Occlusion Field}\label{sec:prior_extraction}
In complex driving scenarios, ego-vehicle blind spots are predominantly caused by the occlusion of dynamic objects like preceding vehicles or pedestrians. Rather than remaining static, these blind spots evolve continuously alongside moving objects, occluding new areas while exposing previously hidden regions. Consequently, effectively repairing these blind spots using collaborative views requires reasoning about the entire spatio-temporal footprint of the occlusion. Motivated by this, we propose an ego-centric Causal Occlusion Field (COF) to explicitly model this temporal evolution and derive occlusion uncertainty as a structured spatial prior.

To achieve this, we first estimate a base dynamic probability map $S_{dyn, k}^\tau$ to identify the current spatial footprint of movable entities. Operating on the shallow, topology-preserving DINO features $\mathbf{F}_{dino, k}^\tau$ alongside the input images, we utilize a DPT-style \citep{Ranftl2020} dynamic head $\Phi_{dyn}$:
\begin{equation}
S_{dyn, k}^\tau = \sigma\left( \Phi_{dyn}(\mathbf{F}_{dino, k}^\tau, \mathbf{I}_k^\tau) \right) \in [0, 1]^{H \times W},
\end{equation}
where $\sigma$ denotes the sigmoid activation. While this static prior isolates the object's current position, it lacks the kinematic context necessary to infer the surrounding occlusion dynamics.
To capture these dynamics, we introduce a temporal-evolutionary occlusion inference mechanism. Operating on the flattened sequence of enhanced features $\tilde{\mathbf{Z}}_{m} \in \mathbb{R}^{B \times (F \cdot L) \times D}$ (where $B$ is the batch size and $F$ is the total number of frames), we apply an \textit{Agent-wise Causal Masked Attention} layer. This mechanism strictly prevents information leakage across different agents or future frames, using a causal binary mask $\mathbf{M}_{cau} \in \mathbb{R}^{(F \cdot L) \times (F \cdot L)}$:
\begin{equation}
\mathbf{M}_{cau}[b, h, i, j] = \begin{cases}
1, & \text{if } a(i) = a(j) \text{ and } \tau(i) \ge \tau(j), \\
0, & \text{otherwise},
\end{cases}
\end{equation}
where $b$ and $h$ index the batch and attention head, respectively, while $i$ and $j$ denote the token indices in the flattened sequence. The functions $a(\cdot)$ and $\tau(\cdot)$ map a given token index to its corresponding agent ID and frame timestamp. Intuitively, this temporal attention acts as a latent tracker. By correlating current features with their historical states within the same agent's perspective, it implicitly extracts a latent kinematic field, formalized as \textit{motion evolution tokens} $\mathbf{Z}_{kin,k}^\tau \in \mathbb{R}^{L \times D}$:
\begin{equation}
\mathbf{Z}_{kin,k}^\tau = \text{Softmax}\left(\frac{\mathbf{Q} \mathbf{K}^T}{\sqrt{d_h}} + \log \mathbf{M}_{cau}\right)\mathbf{V},
\end{equation}
where $\log \mathbf{M}_{cau}$ assigns $-\infty$ to the disallowed entries, strictly enforcing causality.
Based on this latent kinematic field, we employ an Occlusion Decoder $\Phi_{occ}$ (a lightweight fully connected network) to model the occlusion evolution. To ensure stable optimization, we formulate this process as a motion-aware residual dilation. Instead of regressing the occlusion prior from scratch, the network explicitly starts with the base dynamic footprint $S_{dyn,k}^\tau \in [0, 1]^{H \times W}$ and gradually learns to expand it under the guidance of the reshaped kinematic field $\mathbf{Z}_{kin,k}^\tau \in \mathbb{R}^{D \times H \times W}$. Specifically, the decoder predicts a spatio-temporal expansion residual $\Delta S_{kin,k}^\tau$:
\begin{equation}
\Delta S_{kin,k}^\tau = \Phi_{occ}(\mathbf{Z}_{kin,k}^\tau \oplus S_{dyn,k}^\tau) \in [-1, 1]^{H \times W}.
\end{equation}
This residual dictates how the object's footprint adaptively expands along its motion trajectory to cover the dynamically occluded background. We then compute the final \textit{Occlusion Uncertainty Prior} $\mathbf{M}_{occ, k}^\tau$ by applying this learned residual directly to the base footprint:
\begin{equation}
\mathbf{M}_{occ, k}^\tau = \text{Clamp}\Big( S_{dyn,k}^\tau + \Delta S_{kin,k}^\tau, \, 0, \, 1 \Big),
\end{equation}
where the $\text{Clamp}(\cdot)$ function restricts the resulting probability values strictly within the valid range of $[0, 1]$. By employing this residual learning paradigm, the network initially anchors on the physical dynamic footprint and progressively learns the optimal spatial expansion. This process yields a continuous causal occlusion field, a structured spatial prior that preserves the coarse blind-spot location while explicitly injecting kinematic evolution. Ultimately, this field enables the model to adaptively adjust its attention around dynamically evolving occlusions based on cross-agent geometric relationships, actively fetching the most beneficial collaborative features for blind-spot completion.

\subsection{Cross-Agent Latent Residual Denoising}\label{sec:st_alignment}
With the established ego-centric causal occlusion field, we formulate cross-agent feature alignment as a constrained latent residual denoising process\footnote{A detailed theoretical justification is provided in Appendix~\ref{sec:appendix_cald}.}. Specifically, we introduce a Cross-Agent Latent Residual Denoising (CALRD) module to act as a spatial modulator \citep{park2019SPADE}. The intuition is that while collaborative features are essential for completing blind spots, they may introduce uncalibrated geometric drift into visible regions. Therefore, CALRD utilizes the inferred occlusion prior $\mathbf{M}_{occ, k}^\tau \in [0, 1]^{H \times W}$ to explicitly modulate feature supplementation, enabling targeted background completion without overwriting the ego vehicle's reliable local geometry. 

To ensure this residual denoising process remains firmly grounded, we extract the clean ego-only features $\mathbf{F}_{e}^\tau$ as a pure structural reference. We feed the spatialized mixed features $\tilde{\mathbf{F}}_{m, k}^\tau \in \mathbb{R}^{D \times H \times W}$, the occlusion prior $\mathbf{M}_{occ, k}^\tau$, and the reference features $\mathbf{F}_{e}^\tau$ into a dual-branch convolutional architecture. The conditional branch encodes the concatenated spatial prior and reference to extract structural conditions $\mathbf{C}_{pri, k}^\tau \in \mathbb{R}^{D \times H \times W}$:
\begin{equation}
\mathbf{C}_{pri, k}^\tau = \mathcal{Z}_{pri} \Big( \Phi_{con}([\mathbf{M}_{occ, k}^\tau, \mathbf{F}_{e}^\tau]) \Big),
\end{equation}
where $\Phi_{con}$ is a lightweight encoder, and $\mathcal{Z}_{pri}$ is a zero-initialized $1 \times 1$ convolution. The base branch processes the mixed tokens $\tilde{\mathbf{F}}_{m, k}^\tau$ using a base feature encoder $\Phi_{fea}$. The final denoised feature $\hat{\mathbf{F}}_{m, k}^\tau$ is obtained by predicting a residual correction and adding it to the noisy mixed tokens:
\begin{equation}
\hat{\mathbf{F}}_{m, k}^\tau = \tilde{\mathbf{F}}_{m, k}^\tau + \underbrace{\mathcal{Z}_{out} \Big( \Phi_{fea}(\tilde{\mathbf{F}}_{m, k}^\tau) + \mathbf{C}_{pri, k}^\tau \Big)}_{\text{Predicted Residual Correction}},
\end{equation}
where $\mathcal{Z}_{out}$ is another zero-initialized $1 \times 1$ convolution.
This spatial modulation acts as a surgical residual denoising mechanism. Rather than directly predicting absolute target features, the network learns a precise residual correction to filter out uncalibrated collaborative noise. For unoccluded regions ($\mathbf{M}_{occ, k}^\tau \approx 0$), the zero-convolutions naturally suppress collaborative updates, protecting the ego vehicle's confident observations. Conversely, within dynamically shifting blind spots ($\mathbf{M}_{occ, k}^\tau > 0$), the residual update dynamically aggregates structural context to inpaint the missing background.

Finally, the denoised feature map $\hat{\mathbf{F}}_{m, k}^\tau$ is decoded by a specialized Gaussian head $\Phi_{gs}$ to predict the properties of 3D Gaussians (e.g., opacity $\alpha$, color $\mathbf{c}$, scale $\mathbf{s}$, and rotation $\mathbf{q}$) directly along the camera rays. Guided by the decoupled dynamic and static streams, $\Phi_{gs}$ predicts multiple Gaussians $\mathcal{G} = \mathcal{G}_{dyn} \cup \mathcal{G}_{sta}$ distributed along each ray. This disentangles the foreground occluder from the collaboratively inpainted background. Additionally, a parallel sky head $\Phi_{sky}$ extracts environment illumination and distant sky semantics $\mathcal{I}_{sky}$ from the global context. During rasterization for novel view synthesis from an arbitrary ego viewpoint $\mathbf{P}_{novel}$, we can selectively render the full scene $\mathcal{I}_{full} = \text{Render}(\mathcal{G}, \mathbf{P}_{novel}) + \mathcal{I}_{sky}$ or seamlessly reveal the underlying static background by explicitly dropping the dynamic Gaussians $\mathcal{I}_{bg} = \text{Render}(\mathcal{G}_{sta}, \mathbf{P}_{novel}) + \mathcal{I}_{sky}$.

\subsection{Model Training}\label{sec:training}
Training a highly dynamic distributed network from scratch is notoriously unstable. To ensure stable convergence, we propose a two-stage progressive training curriculum.

\textbf{Stage I: Single-Agent Pre-training.} We first establish robust ego-centric reconstruction and kinematic priors. With the base VGGT backbone kept frozen, the Gaussian rendering head $\Phi_{gs}$, sky head $\Phi_{sky}$, and dynamic head $\Phi_{dyn}$ are trained using exclusively single-agent sequences. The network minimizes photometric losses $\mathcal{L}_{pho}$ for scene rendering, which calculates the discrepancy between the synthesized image $\mathcal{I}_{full}$ and the ground-truth image $\mathcal{I}_{gt}$ using a combination of $L_1$ and LPIPS \citep{zhang2018perceptual} distances:
\begin{equation}
\mathcal{L}_{pho} = \lambda_1 \Vert \mathcal{I}_{full} - \mathcal{I}_{gt} \Vert_1 + \lambda_{lpips} \text{LPIPS}(\mathcal{I}_{full}, \mathcal{I}_{gt}).
\end{equation}
Simultaneously, $\Phi_{dyn}$ is supervised via Binary Cross-Entropy (BCE) using pseudo ground-truth dynamic masks $S_{msk}^{gt}$ extracted by a pre-trained SegFormer \citep{xie2021segformer}:
\begin{equation}
\mathcal{L}_{msk} = -\frac{1}{|\Omega|} \sum_{\mathbf{u} \in \Omega} \Big[ S_{msk}^{gt}(\mathbf{u}) \log S_{dyn,k}^\tau(\mathbf{u}) + \big(1 - S_{msk}^{gt}(\mathbf{u})\big) \log \big(1 - S_{dyn,k}^\tau(\mathbf{u})\big) \Big],
\end{equation}
where $\mathbf{u}$ denotes the 2D pixel coordinates and $\Omega$ represents the entire spatial image domain. The total loss for Stage I is $\mathcal{L}_{s1} = \mathcal{L}_{pho} + \lambda_{msk} \mathcal{L}_{msk}$.

\textbf{Stage II: Cross-Agent Adaptation.} Initialized with Stage I weights, we further freeze the pre-trained dynamic head $\Phi_{dyn}$, sky head $\Phi_{sky}$, and train the newly added \texttt{FRUC} modules alongside other trainable decoders in stage I using multi-agent sequences. To explicitly guide the residual learning, we introduce a set of auxiliary objectives $\mathcal{L}_{aux}$ (detailed in Appendix~\ref{sec:appendix_impl}). These losses establish a zero-sum dynamic: anchoring the occlusion prior to prevent unbounded expansion, driving background completion within occlusion boundaries, and enforcing global latent denoising to preserve the reliable ego-only baseline. Stage II combines these penalties with the base photometric loss:
\begin{equation}
\mathcal{L}_{s2} = \mathcal{L}_{pho} + \mathcal{L}_{aux}.
\end{equation}
Coupled with the zero-initialization strategy, this joint objective transforms the highly non-convex multi-agent optimization into a bounded, highly stable residual learning problem.

\section{Experiments}\label{sec:experiments}

\subsection{Experimental Setup}

\begin{figure}[t]
\centering
\includegraphics[width=\linewidth]{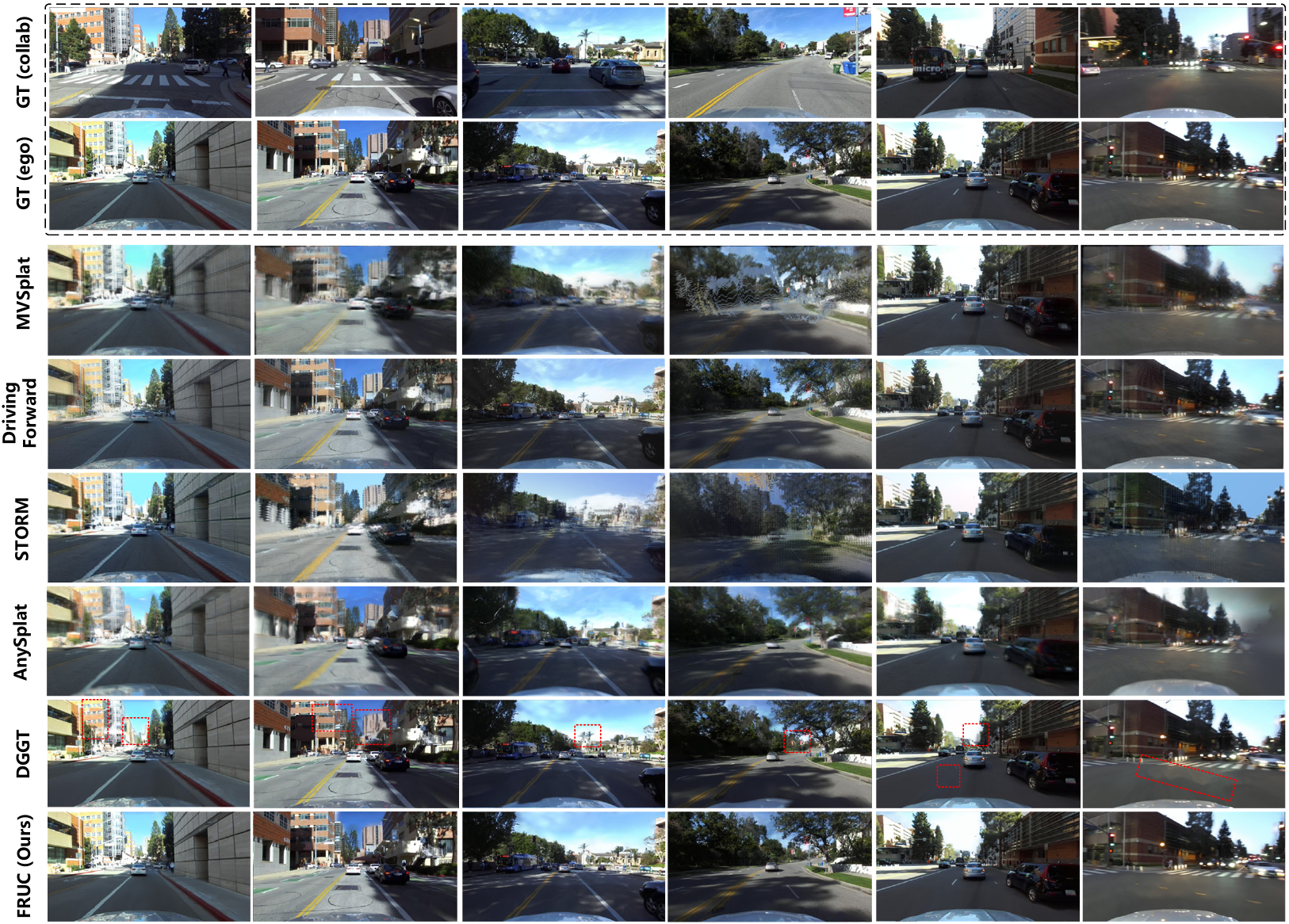}
\vspace{-6mm}
\caption{\textbf{Qualitative comparison of NVS on V2X-Real.} The input context consists of both ego and collaborative views at $t\!-\!1$ and $t\!+\!1$, and the task is to reconstruct the ego view at time $t$. Here we show the results of forward-facing camera. \textcolor{red}{Red} boxes highlight degradations of second-best baseline.}
\vspace{-2mm}
\label{fig:nvs_qualitative}
\end{figure}

\textbf{Datasets.} 
We conduct primary benchmark evaluations on V2X-Real \citep{xiangv2xreal} and evaluate generalizability on UrbanIng-V2X dataset \citep{sekaran2025urbaningvx}. To ensure high-quality and meaningful evaluation, we carefully pre-process the raw datasets. This involves supplementing the missing dynamic and sky masks, filtering out invalid sequences, and explicitly designating the ego vehicle, ultimately establishing two rigorous benchmarks tailored for collaborative dynamic scene reconstruction. Detailed data processing pipelines and filtering criteria are provided in Appendix~\ref{sec:appendix_impl}.

\noindent \textbf{Implementation Details and Baselines.} 
Our model builds upon the VGGT-1B \citep{wang2025vggt} backbone and is trained on 4 NVIDIA RTX 5880 GPUs. The image resolution is set to $378 \times 672$ and the models are trained for 10 epochs.
We compare our model against scene-optimized techniques like EmerNeRF \citep{yang2023emernerf}, 3DGS \citep{kerbl20233d}, and V2X-Gaussians \citep{11097436}, as well as feed-forward methods like MVSplat \citep{chen2024mvsplat}, DrivingForward \citep{tian2025drivingforward}, STORM \citep{yang2024stormspatiotemporalreconstructionmodel}, AnySplat \citep{jiang2025anysplat}, and DGGT \citep{chen2025dggtfeedforward4dreconstruction}. Except for V2X-Gaussians, all baselines are single-agent methods. For fair comparison, we adapt them by simply replacing their original single-agent sequence inputs with multi-agent sequences, while keeping their model architectures and training methods entirely unchanged. This adaptation is justified since collaborative reconstruction can be reformulated as a generalized single-agent reconstruction problem under a dynamic multi-camera setup. More details are in Appendix~\ref{sec:appendix_impl}.

\noindent \textbf{Evaluation Metrics.}
To comprehensively assess reconstruction quality, we report PSNR, SSIM, and LPIPS for full image and dynamic-only settings, while using NIQE specifically to evaluate the perceptual realism of blind-spot completion. We default to the multi-frame (MF) mode for our primary benchmark comparisons, and provide additional quantitative results under the single-frame (SF) mode in Appendix~\ref{sec:appendix_more_results}, along with detailed metric formulations.

\subsection{Benchmark Comparison}

\begin{table*}[t]
\centering
\caption{\textbf{Comparison to state-of-the-art methods on the V2X-Real Dataset.} We input two frames of multi-agent views to reconstruct the intermediate ego view. \textbf{Best} in bold, \underline{second-best} underlined.}
\label{tab:main_results}
\resizebox{\textwidth}{!}{
\small
\renewcommand{\arraystretch}{0.8}
\begin{tabular}{lcccccccc}
\toprule
\multirow{2}{*}{\textbf{Methods}} & \multicolumn{3}{c}{\textbf{Full image}} & \multicolumn{3}{c}{\textbf{Dynamic-only}} & \multirow{2}{*}{\shortstack{\textbf{Blind-Spot}\\\textbf{(NIQE$\downarrow$)}}} & \multirow{2}{*}{\shortstack{\textbf{Inference}\\\textbf{Time $\downarrow$}}} \\
\cmidrule(lr){2-4} \cmidrule(lr){5-7}
& \textbf{PSNR$\uparrow$} & \textbf{SSIM$\uparrow$} & \textbf{LPIPS$\downarrow$} & \textbf{PSNR$\uparrow$} & \textbf{SSIM$\uparrow$} & \textbf{LPIPS$\downarrow$} & & \\
\midrule
\multicolumn{9}{l}{\textit{Per-scene Optimization Methods}} \\
EmerNeRF \citep{yang2023emernerf} & 21.65 & 0.636 & 0.583 & 17.79 & 0.979 & 0.043 & - & 10min \\
3DGS \citep{kerbl20233d} & 20.19 & 0.616 & 0.592 & 16.85 & 0.976 & 0.035 & - & 14min \\
V2X-Gaussians \citep{11097436} & \underline{22.20} & 0.784 & 0.409 & 23.87 & \textbf{0.994} & \underline{0.011} & - & 21min \\
\midrule
\multicolumn{9}{l}{\textit{Generalizable Feed-forward Methods}} \\
MVSplat \citep{chen2024mvsplat} & 16.73 & 0.478 & 0.281 &  15.18 & 0.952 & 0.049 & - & 0.31s \\
DrivingForward \citep{tian2025drivingforward} & 17.37 & 0.596 & 0.239 & 14.46 & 0.955 & 0.047 & - & \textbf{0.15s} \\
STORM \citep{yang2024stormspatiotemporalreconstructionmodel} & 16.27 & 0.579 & 0.379 & 14.36 & 0.953 & 0.069 & 8.682 & 0.35s \\
AnySplat \citep{jiang2025anysplat} & 19.18 & 0.605 & 0.219 & 16.69 & 0.959 & 0.046 & - & \underline{0.29s} \\
DGGT \citep{chen2025dggtfeedforward4dreconstruction} & 21.85 & \underline{0.804} & \underline{0.129} & \underline{24.76} & \underline{0.992} & 0.129 & \underline{4.301} & 0.54s \\
\midrule
\textbf{\texttt{FRUC}} (Ours) & \textbf{25.89} & \textbf{0.883} & \textbf{0.076} & \textbf{25.93} & \textbf{0.994} & \textbf{0.007} & \textbf{3.949} & 0.77s \\
\bottomrule
\vspace{-8mm}
\end{tabular}
}
\end{table*}

\begin{table*}[t]
\centering
\caption{\textbf{Generalizability Comparison on the UrbanIng-V2X Dataset.} We evaluate NVS quality under both zero-shot and trained settings. \textbf{Best} in bold, and \underline{second-best} underlined.}
\label{tab:zero_shot_urbaning}
\resizebox{\textwidth}{!}{
\footnotesize
\renewcommand{\arraystretch}{0.8}
\begin{tabular}{lccccccc}
\toprule
\multirow{2}{*}{\textbf{Method}} & \multicolumn{3}{c}{\textbf{Full image}} & \multicolumn{3}{c}{\textbf{Dynamic-only}} & \multirow{2}{*}{\shortstack{\textbf{Blind-Spot}\\\textbf{(NIQE$\downarrow$)}}} \\
\cmidrule(lr){2-4} \cmidrule(lr){5-7}
& \textbf{PSNR$\uparrow$} & \textbf{SSIM$\uparrow$} & \textbf{LPIPS$\downarrow$} & \textbf{PSNR$\uparrow$} & \textbf{SSIM$\uparrow$} & \textbf{LPIPS$\downarrow$} & \\
\midrule
\multicolumn{8}{l}{\textit{Zero-shot} \textit{(Trained on V2X-Real)}} \\
MVSplat \citep{chen2024mvsplat} & 18.58 & 0.463 & 0.380 & 15.15 & 0.966 & 0.056 & - \\
DrivingForward \citep{tian2025drivingforward} & 18.09 & 0.676 & 0.335 & 15.66 & 0.970 & 0.037 & - \\
STORM \citep{yang2024stormspatiotemporalreconstructionmodel} & 17.30 & 0.604 & 0.458 & 14.16 & 0.967 & 0.068 & 11.781 \\
AnySplat \citep{jiang2025anysplat} & 21.31 & 0.552 & \underline{0.222} & 17.88 & 0.974 & \underline{0.028} & - \\
DGGT \citep{chen2025dggtfeedforward4dreconstruction} & \underline{22.23} & \underline{0.724} & 0.286 & \underline{23.71} & \textbf{0.993} & 0.208 & \underline{7.509} \\
\texttt{FRUC} (Ours) & \textbf{25.07} & \textbf{0.795} & \textbf{0.144} & \textbf{24.61} & \textbf{0.993} & \textbf{0.008} & \textbf{5.270} \\
\midrule
\multicolumn{8}{l}{\textit{Trained} \textit{(Fine-tuned on UrbanIng-V2X)}} \\
MVSplat \citep{chen2024mvsplat} & 19.24 & 0.467 & 0.345 & 15.86 & 0.969 & 0.039 & - \\
DrivingForward \citep{tian2025drivingforward} & 21.80 & \underline{0.754} & 0.258 & 16.71 & 0.974 & \underline{0.028} & - \\
STORM \citep{yang2024stormspatiotemporalreconstructionmodel} & 17.36 & 0.614 & 0.336 & 14.22 & 0.969 & 0.079 & 6.268 \\
AnySplat \citep{jiang2025anysplat} & 22.45 & 0.579 & \underline{0.176} & 19.20 & \underline{0.976} & \textbf{0.025} & - \\
DGGT \citep{chen2025dggtfeedforward4dreconstruction} & \underline{25.37} & 0.745 & 0.275 & \underline{24.63} & \textbf{0.993} & 0.285 & \underline{5.728} \\
\texttt{FRUC} (Ours) & \textbf{27.26} & \textbf{0.996} & \textbf{0.005} & \textbf{27.64} & 0.819 & 0.079 & \textbf{3.854} \\
\bottomrule
\vspace{-7mm}
\end{tabular}
}
\end{table*}

\textbf{Novel View Synthesis (NVS).} We evaluate the rendering quality from the ego vehicle's perspective. In the default MF setting, the input context consists of 4 uncalibrated images derived from a specific ego-collab view pair across two consecutive timestamps (i.e., ego and collaborative views at $t\!-\!1$ and $t\!+\!1$) to reconstruct the target ego image at the intermediate time $t$. Quantitative results are averaged across all potential cross-agent view pairs that possess semantic co-visibility. Table~\ref{tab:main_results} reports the comprehensive benchmark, where \texttt{FRUC} outperforms all baseline methods by a large margin. Notably, the degraded performance of DGGT and AnySplat indicates that naive multi-agent feature aggregation struggles with cross-agent geometric misalignment. Furthermore, despite incorporating advanced spatial reasoning modules to handle uncalibrated collaborative context, our framework maintains a highly competitive inference speed of $0.77s$, significantly faster than optimization-based paradigms. Fig.~\ref{fig:nvs_qualitative} further provides qualitative comparisons. While baseline methods often exhibit severe artifacts or fail to maintain the ego vehicle's correct spatial layout, our model successfully resolves the geometric interference and produces the most faithful and detailed scene reconstructions.

\textbf{Generalizability.}
We evaluate the generalization capability of our model on the UrbanIng-V2X dataset. As shown in Table~\ref{tab:zero_shot_urbaning}, we conduct evaluations under both zero-shot (direct inference using weights trained on V2X-Real) and trained (fine-tuning on UrbanIng-V2X) settings. In the zero-shot scenario, FRUC consistently outperforms feed-forward baselines, demonstrating robust cross-dataset adaptability and reliable structural priors learned from V2X-Real. When fine-tuned on the new dataset, our model further solidifies its advantage across most metrics, indicating strong capacity to fit new collaborative environments and diverse driving scenarios.
\begin{figure}[t]
\centering
\includegraphics[width=\linewidth]{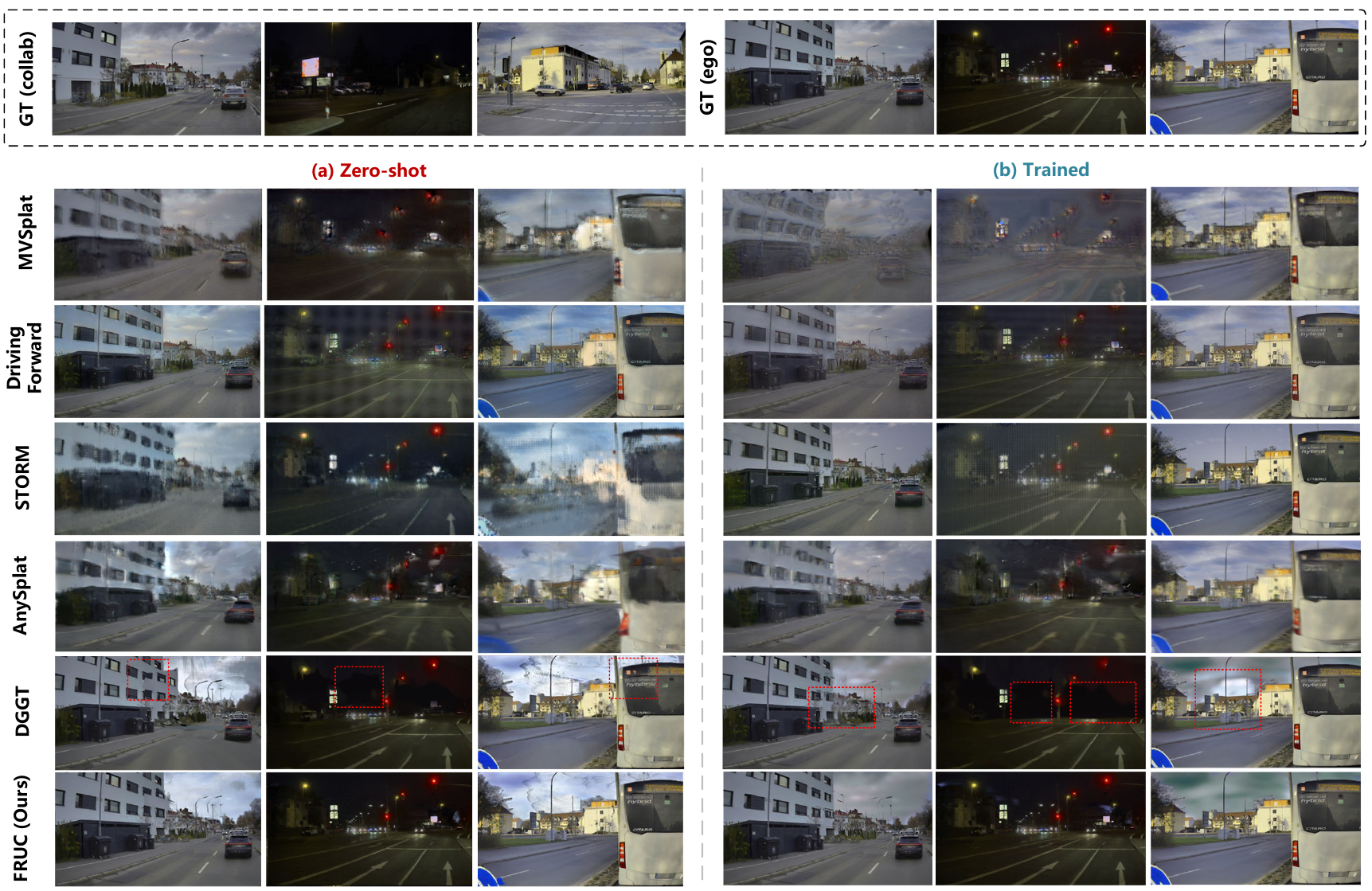}
\vspace{-6mm}
\caption{\textbf{Qualitative comparison of NVS on UrbanIng-V2X.} We compare the zero-shot and trained results under MF settings. \textcolor{red}{Red} boxes highlight degradations of second-best baseline.}
\vspace{-4mm}
\label{fig:nvs_qualitative_urban}
\end{figure}

Fig.~\ref{fig:nvs_qualitative_urban} further provides qualitative comparisons on UrbanIng-V2X under both settings. For \textit{zero-shot}, \texttt{FRUC} transfers well from V2X-Real to UrbanIng-V2X, producing substantially clearer scene layouts and more stable object boundaries than the competing feed-forward baselines, which often suffer from severe blur, ghosting, or structural collapse. In the \textit{trained} setting, this advantage becomes more pronounced: \texttt{FRUC} further improves local sharpness, lighting consistency, and geometric integrity, while maintaining reliable reconstruction of distant traffic lights, building facades, and large dynamic objects. These qualitative results corroborate the quantitative improvements, demonstrating robust cross-dataset generalization and strong adaptation after fine-tuning.
More results are in Appendix \ref{sec:appendix_more_results}.

\begin{wrapfigure}{r}{0.5\textwidth}
\vspace{-6mm}
\centering
\includegraphics[width=\linewidth]{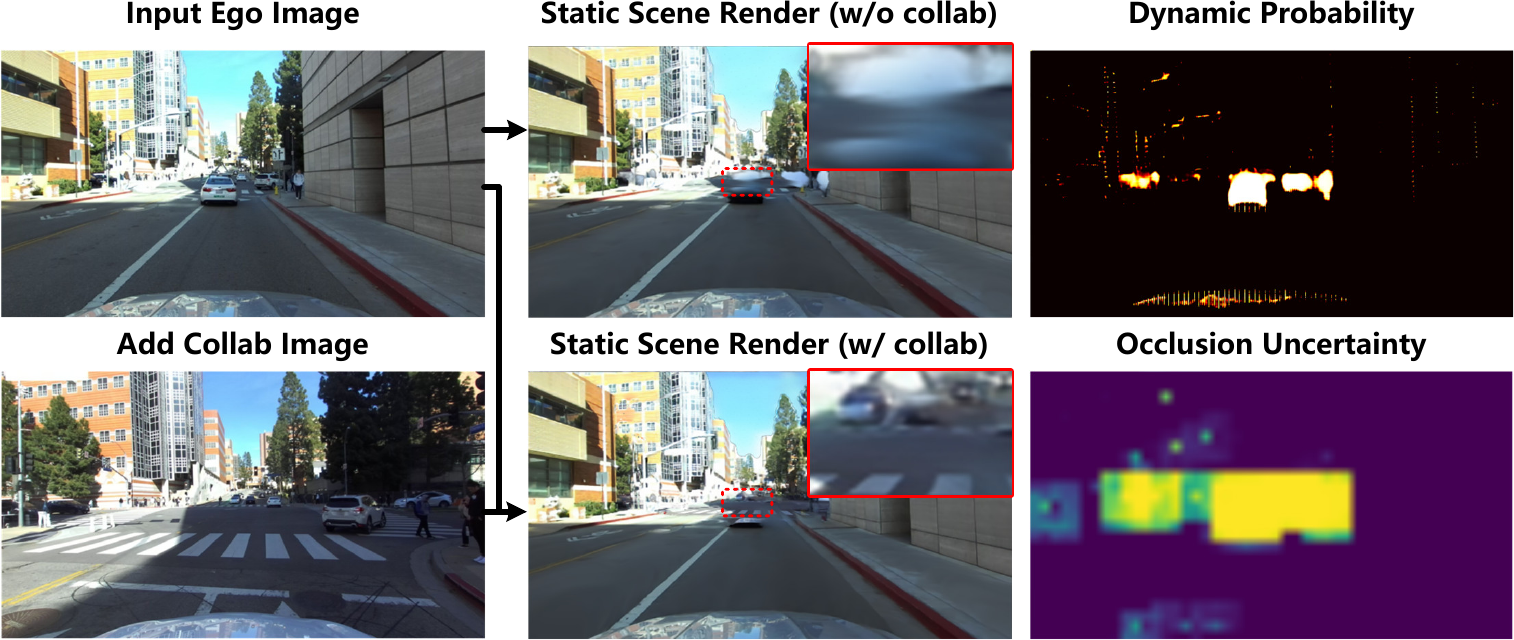}
\vspace{-6mm}
\caption{\textbf{Blind-spot completion on V2X-Real.}}
\vspace{-4mm}
\label{fig:blindspot_qualitative}
\end{wrapfigure}

\textbf{Blind-Spot Completion and Scene Editing.} High-fidelity blind-spot completion is essential for reliable scene editing. To quantify this, we render the static scene and compute NIQE exclusively on cropped regions around the original dynamic entities. Tables~\ref{tab:main_results} and~\ref{tab:zero_shot_urbaning} show that \texttt{FRUC} achieves the best NIQE across both datasets, demonstrating superior blind-spot completion capability. Fig.~\ref{fig:blindspot_qualitative} further compares ego-only and collaborative reconstruction. While ego-only reconstruction reveals clear geometric holes and blurring after static rendering, \texttt{FRUC} leverages collaborative views to seamlessly inpaint the missing background, exposing a realistic blind-spot view and preserving reliable ego-observed regions. 

\begin{figure}[t]
\centering
\includegraphics[width=\linewidth]{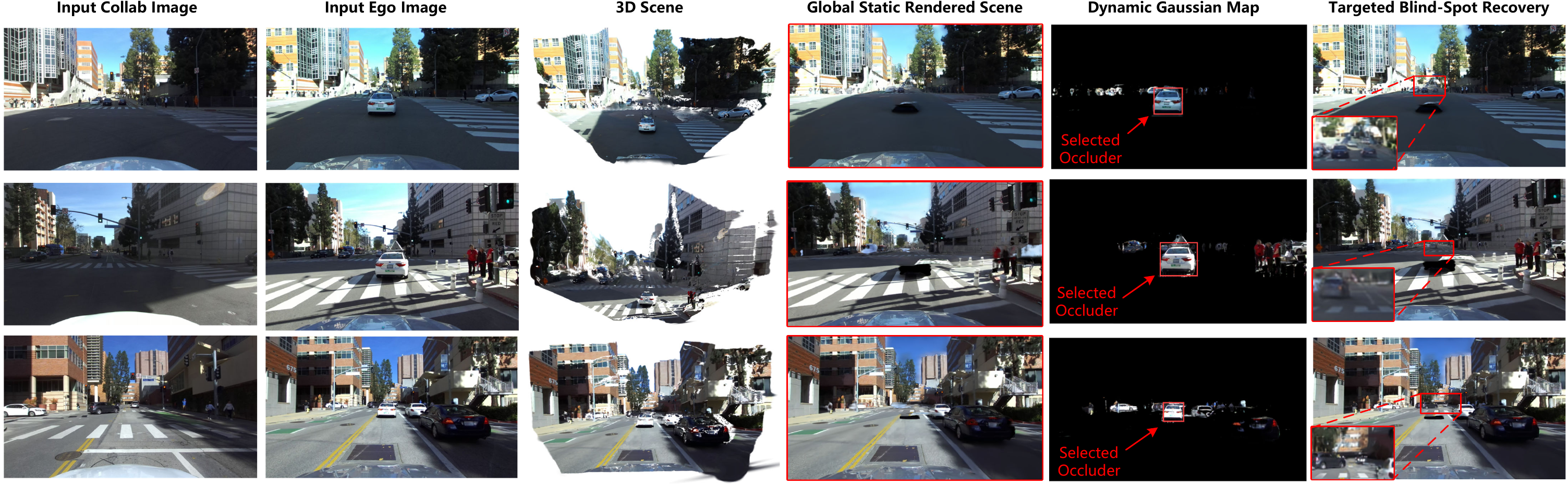}
\vspace{-6mm}
\caption{\textbf{More Scene Editing Results on V2X-Real.} We explicitly disentangle the 3D scene into global static and dynamic representations. By identifying specific dynamic Gaussians, our framework enables targeted blind-spot recovery utilizing collaborative context.}
\vspace{-4mm}
\label{fig:vis_editing}
\end{figure}

Fig.~\ref{fig:vis_editing} demonstrates how this capability supports object-level scene editing. Since \texttt{FRUC} explicitly decouples the reconstructed 3D space into static and dynamic representations, it naturally supports flexible rendering manipulations. Beyond simply rendering the global scene, our method enables fine-grained, object-level editing. As shown in the figure, by isolating specific dynamic Gaussians corresponding to a target occluder, we can selectively remove it from the scene. The framework then seamlessly fills the resulting "hole" by rendering the geometrically consistent static background, successfully achieving targeted blind-spot recovery utilizing the integrated collaborative context.

\subsection{Ablation Studies}

\begin{table*}[t]
\centering
\caption{\textbf{Ablation Studies on the V2X-Real Dataset.}  \textbf{Best} in bold, and \underline{second-best} underlined.}
\label{tab:ablation}
\newcommand{\gcheck}{\textcolor{green!60!black}{\ding{51}}}
\newcommand{\rcross}{\textcolor{red!75!black}{\ding{55}}}
\resizebox{\textwidth}{!}{
\footnotesize
\renewcommand{\arraystretch}{0.8}
\begin{tabular}{ccc|ccc|ccc|c}
\toprule
\multicolumn{3}{c|}{\textbf{Components}} & \multicolumn{3}{c|}{\textbf{Full Image}} & \multicolumn{3}{c|}{\textbf{Dynamic-only}} & \multirow{2}{*}{\shortstack{\textbf{Blind-Spot}\\\textbf{(NIQE$\downarrow$)}}} \\
\cmidrule(lr){1-3} \cmidrule(lr){4-6} \cmidrule(lr){7-9}
\textbf{CALRD} & \textbf{COF} & \textbf{Aux.} & \textbf{PSNR$\uparrow$} & \textbf{SSIM$\uparrow$} & \textbf{LPIPS$\downarrow$} & \textbf{PSNR$\uparrow$} & \textbf{SSIM$\uparrow$} & \textbf{LPIPS$\downarrow$} & \\
\midrule
\rcross & \rcross & \gcheck & 21.94 & 0.799 & 0.138 & 24.71 & 0.985 & 0.098 & 4.181 \\
\gcheck & \rcross & \gcheck & \underline{23.97} & \underline{0.836} & \underline{0.119} & \underline{24.96} & \underline{0.989} & \underline{0.077} & \underline{4.017} \\
\gcheck & \gcheck & \rcross & 22.76 & 0.801 & 0.121 & 24.92 & 0.987 & 0.084 & 4.309 \\
\gcheck & \gcheck & \gcheck & \textbf{25.89} & \textbf{0.883} & \textbf{0.076} & \textbf{25.93} & \textbf{0.994} & \textbf{0.007} & \textbf{3.949} \\
\bottomrule
\end{tabular}
}
\vspace{-4mm}
\end{table*}

\textbf{Effectiveness of Key Components.} 
Table~\ref{tab:ablation} show that CALRD and COF are the two core components. When both modules are removed, the performance drops most severely, indicating that naive multi-agent feature fusion cannot reliably resolve cross-agent geometric interference. When CALRD is preserved but COF is removed, the performance recovers only partially, which suggests that residual denoising alone is insufficient without a structured causal prior telling the model where collaborative completion is needed. In essence, COF provides the spatially localized, motion-aware occlusion guidance, while CALRD converts this guidance into stable ego-centric feature correction.

\textbf{Effectiveness of Auxiliary Losses.} 
Removing the auxiliary losses while keeping both CALRD and COF still produces a functional model, but the performance drops clearly, especially in blind-spot completion. This indicates that $\mathcal{L}_{\text{aux}}$ is not the source of the architectural gain itself. Instead, it serves as an optimization stabilizer that regularizes the occlusion prior, constrains the residual denoising trajectory, and encourages cooperative completion to evolve in a physically meaningful direction.

\section{Conclusion}\label{sec:conclusion}
\noindent In this paper, we have introduced \texttt{FRUC}, a feed-forward 3DGS framework for dynamic scene reconstruction from uncalibrated collaborative driving views. 
To bypass the strict requirement for precise cross-agent calibration, \texttt{FRUC} elegantly re-formulates distributed multi-agent inputs as a spatio-temporally unstructured ego-centric multi-camera system. 
Specifically, we have proposed an ego-centric causal occlusion field to extract motion-aware spatial priors, and framed cross-agent feature fusion as a deterministic latent residual denoising process. This design effectively completes dynamic ego-blind spots without much corrupting reliable local observations, achieving state-of-the-art reconstruction quality and inference efficiency on V2X-Real and UrbanIng-V2X benchmarks.

\noindent \textbf{Limitations and Future Work.} Currently, our method may struggle in severe occlusion scenarios where weak cross-agent semantic correlation limits collaborative benefits. Besides, inaccurate dynamic probability maps can lead to failure cases, compromising the reliability of downstream scene editing. Future work will focus on enhancing the capacity of collaborative reconstruction in such severe occlusion scenarios and improving the robustness of dynamic modeling.

\noindent \textbf{Acknowledgments.} The research work described in this paper was conducted in the JC STEM Lab of Smart City funded by The Hong Kong Jockey Club Charities Trust under Contract 2023-0108. The work was supported in part by the Hong Kong SAR Government under the Global STEM Professorship and Research Talent Hub.

\newpage
\bibliographystyle{plainnat}
\bibliography{egbib}

\newpage
\appendix
\newpage

\appendix

\label{sec:appendix}

\section{More Implementation Details}
\label{sec:appendix_impl}

\subsection{Dataset Preparation and Utilization}\label{sec:appendix_dataset}

\textbf{Data Preparation.} We unify V2X-Real \citep{xiangv2xreal} and UrbanIng-V2X \citep{sekaran2025urbaningvx} into a common OPV2V \citep{xu2022opencood} format benchmark. The processed benchmark preserves the full original per-agent data layout, including RGB images, frame ids, camera parameters, ego poses, scene context files, and LiDAR files. On top of this structure, we generate sky masks and fine dynamic masks using an off-the-shelf SegFormer \citep{xie2021segformer} trained on Cityscapes classes. We also establish cross-agent view-association metadata by coarsely estimating spatial overlap via LiDAR coordinates, camera poses, and Field of View (FOV) intersections, followed by manual verification. This metadata is specifically curated to filter out distant or non-overlapping ego-collaborative view pairs, ensuring the construction of a high-quality evaluation set. For UrbanIng-V2X, we additionally convert the raw multi-vehicle logs into the same file layout, remap camera ids to match the V2X-Real convention, and apply the same split protocol. Table~\ref{tab:dataset_statistics} summarizes the resulting benchmark structure and clarifies the split sizes.

\textbf{Data Utilization.} For model inference, our pipeline strictly utilizes \textbf{only} RGB images and frame ids, which serve as the primary visual and temporal inputs, and all other signals are excluded from the model inputs. The sky masks and dynamic masks act as auxiliary supervision rather than inputs. The dynamic masks provide ground-truth supervision for the dynamic-head during training and are repurposed solely for masked metric computation during evaluation, while the sky masks assist training supervision and metric computation. To strictly preserve our fully uncalibrated testing paradigm, any global pose, LiDAR information, or view-association metadata is explicitly excluded from the model's inference inputs. Instead, the view-association metadata acts purely as an offline oracle exclusively during evaluation to construct valid collaborative benchmark pairs and determine the relevant collaborative view context for the ego vehicle. In this way, all auxiliary signals are confined to training supervision, benchmark construction, and metric computation, leaving the model to run on uncalibrated RGB images alone.

\subsection{Model Training Details}\label{sec:appendix_loss}

During the Stage II Cross-Agent Adaptation, we explicitly supervise the model training with auxiliary losses. In real-world collaborative driving, blindly fusing multi-agent features often leads to geometric contamination and severe artifact generation due to uncalibrated errors. To mitigate this, our auxiliary losses are designed to establish an information-theoretic balance between single-agent reliability and collaborative completion. We categorize these auxiliary objectives into three functions: \textit{Prior Stabilization}, \textit{Cooperative Completion}, and \textit{Ego-Manifold Regularization}.

\ding{182} \textbf{Prior Stabilization ($\mathcal{L}_{reg}$ and $\mathcal{L}_{var}$).}
The occlusion uncertainty prior $\mathbf{M}_{occ, k}^\tau$ dictates the routing of collaborative features. To prevent this prior from either expanding unbounded or collapsing to zero, we introduce two regularizers. First, the \textit{Prior Consistency Loss} $\mathcal{L}_{reg}$ anchors the inferred occlusion mask to a dilated base dynamic footprint $\tilde{S}_{dyn, k}^\tau = \mathcal{D}_{\mathcal{K}}(S_{dyn, k}^\tau)$ using an L1 penalty:
\begin{equation}
\mathcal{L}_{reg} = \Vert \mathbf{M}_{occ, k}^\tau - \tilde{S}_{dyn, k}^\tau \Vert_1.
\end{equation}
Here, $\mathcal{D}_{\mathcal{K}}(\cdot)$ denotes the morphological dilation operator with a 2D square structuring element (kernel) $\mathcal{K}$ of a predefined size (e.g., $15 \times 15$ pixels). This operation expands the boundaries of the original dynamic mask $S_{dyn, k}^\tau$ to encompass its immediate surrounding context. This instills a conservative principle: the ego vehicle should fundamentally trust its local high-fidelity observations for static, visible regions, reserving collaborative feature aggregation strictly for dynamically occluded regions. Second, to encourage spatial diversity within the inferred mask and prevent constant-state collapse, we introduce a \textit{Prior Dispersion Loss} $\mathcal{L}_{var}$:
\begin{equation}
\mathcal{L}_{var} = -\text{Var}(\mathbf{M}_{occ, k}^\tau),
\end{equation}
where $\text{Var}(\cdot)$ denotes the spatial variance.

\ding{183} \textbf{Cooperative Completion ($\mathcal{L}_{coo}$).}
Reconstructing occluded backgrounds behind moving objects is an inherently ill-posed inverse problem if relying solely on single-agent temporal priors. To provide the explicit driving force for cross-view geometric completion, we design the \textit{Cooperative Completion Loss} $\mathcal{L}_{coo}$. Since dynamic objects cast the most severe visual shadows, we define an active optimization target exclusively around their immediate boundaries. Specifically, we extract the dynamic occlusion ring (or shadow boundary) $S_{bnd, k}^\tau$ by subtracting the base dynamic footprint from its dilated version: $S_{bnd, k}^\tau = \mathcal{D}_{\mathcal{K}}(S_{dyn, k}^\tau) \setminus S_{dyn, k}^\tau$. During forward rendering, we synthesize a background-only image $\mathcal{I}_{bg, k}^\tau$ by explicitly omitting dynamic Gaussians. By enforcing a photometric penalty between $\mathcal{I}_{bg, k}^\tau$ and the ground-truth image $\mathcal{I}_{gt, k}^\tau$ exclusively within this occlusion boundary $S_{bnd, k}^\tau$, we create an intense economic optimization pressure:
\begin{equation}
\mathcal{L}_{coo} = \frac{1}{|S_{bnd, k}^\tau|} \sum_{\mathbf{u} \in S_{bnd, k}^\tau} \Vert \mathcal{I}_{bg, k}^\tau(\mathbf{u}) - \mathcal{I}_{gt, k}^\tau(\mathbf{u}) \Vert_1,
\end{equation}
where $|S_{bnd, k}^\tau|$ is the total number of valid pixels in the occlusion boundary mask. This forces the network to expand $\mathbf{M}_{occ, k}^\tau$ to fetch and align features from collaborative agents to reconstruct these hidden geometries.

\begin{table*}[t]
\newcommand{\gcheck}{\textcolor{green!60!black}{\ding{51}}}
\newcommand{\rcross}{\textcolor{red!75!black}{\ding{55}}}
\centering
\caption{\textbf{Processed Dataset Summary.} Green checkmarks (\gcheck) denote that the specific data modality is exposed to the model pipeline during the corresponding phase, whereas red crosses (\rcross) indicate it is strictly excluded. The sample counts represent the total number of distinct ego-collaborative view pairs in the dataset, reported in thousands (K).}
\label{tab:dataset_statistics}
\footnotesize
\renewcommand{\arraystretch}{0.9}
\resizebox{\textwidth}{!}{
\begin{tabular}{ll|cccccc|ccc|c}
\toprule
\multirow{2}{*}{\shortstack[c]{\textbf{Dataset}}} & \multirow{2}{*}{\shortstack[c]{\textbf{Split}}} & \multicolumn{6}{c|}{\textbf{Original Data}} & \multicolumn{3}{c|}{\textbf{Added / Parsed Metadata}} & \multirow{2}{*}{\shortstack[c]{\textbf{\# Samples}\\\textbf{(K)}}} \\
\cmidrule(lr){3-8} \cmidrule(lr){9-11}
& & \shortstack{\textbf{RGB}\\\textbf{Images}} & \shortstack{\textbf{Frame}\\\textbf{ID}} & \shortstack{\textbf{Camera}\\\textbf{Params}} & \shortstack{\textbf{Ego}\\\textbf{Pose}} & \shortstack{\textbf{Scene}\\\textbf{Context}} & \shortstack{\textbf{LiDAR}\\\textbf{Info}} & \shortstack{\textbf{Sky}\\\textbf{Mask}} & \shortstack{\textbf{Dynamic}\\\textbf{Mask}} & \shortstack{\textbf{View}\\\textbf{Association}} & \\
\midrule
\multirow{2}{*}{V2X-Real} & Train & \gcheck & \gcheck & \rcross & \rcross & \rcross & \rcross & \gcheck & \gcheck & \rcross & 103.5 \\
& Val & \gcheck & \gcheck & \rcross & \rcross & \rcross & \rcross & \gcheck & \gcheck & \gcheck & 29.1 \\
\midrule
\multirow{2}{*}{UrbanIng-V2X} & Train & \gcheck & \gcheck & \rcross & \rcross & \rcross & \rcross & \gcheck & \gcheck & \rcross & 114.5 \\
& Val & \gcheck & \gcheck & \rcross & \rcross & \rcross & \rcross & \gcheck & \gcheck & \gcheck & 44.6 \\
\bottomrule
\vspace{-6mm}
\end{tabular}
}
\end{table*}

\ding{184} \textbf{Ego-Manifold Regularization ($\mathcal{L}_{den}$).}
Finally, to prevent catastrophic forgetting of the ego-only capabilities during multi-agent fine-tuning, we employ a \textit{Ego-Reference Denoising Loss} $\mathcal{L}_{den}$. Inspired by latent denoising \citep{ho2020denoisingdiffusionprobabilisticmodels}, it applies a global Mean Squared Error (MSE) objective on the latent space representations:
\begin{equation}
\mathcal{L}_{den} = \big\| \hat{\mathbf{F}}_{m, k}^\tau - \mathbf{F}_{e}^\tau \big\|_2^2.
\end{equation}
Crucially, this loss is applied \textit{globally} without spatial masking. This establishes a competitive dynamic: the global MSE acts as a strong structural prior pulling the entire feature map towards the safe ego-only baseline; meanwhile, the cooperative completion loss $\mathcal{L}_{coo}$ provides a counter-gradient, forcing the network to deviate from the ego baseline \textit{only} in severely occluded regions where the collaborative features offer a significant rendering advantage that outweighs the MSE penalty. Additionally, to ensure ego-centric structural dominance during optimization, we employ an asymmetric loss weighting strategy that more prioritizes the ego vehicle's views (weight $1.0$) over collaborative views (weight $0.1$) across all photometric and perceptual penalties.

\subsection{Baseline Implementation Details}\label{sec:appendix_baselines}

Except for V2X-Gaussians \citep{11097436}, all comparison methods are originally designed for single-vehicle inputs. To ensure a fair comparison, we adapt them to the cooperative setting only through data reorganization, while keeping their backbone architectures, Gaussian prediction heads, rendering pipelines, and original training objectives unchanged. Concretely, each cooperative sample is converted into a unified multi-view temporal input centered on the ego agent, so that all methods receive the same cross-agent observations under a shared ego-centric reference frame.

For feed-forward multi-view methods such as AnySplat \citep{jiang2025anysplat}, MVSplat \citep{chen2024mvsplat}, and DGGT \citep{chen2025dggtfeedforward4dreconstruction}, the adaptation is limited to replacing the original single-vehicle data loader with a cooperative multi-view input interface. The source and target views are reorganized to match each method's expected input format, but the encoder-decoder structure, cost-volume construction, Gaussian decoding, and loss design remain exactly the same as in the original implementations. For temporal reconstruction methods such as STORM \citep{yang2024stormspatiotemporalreconstructionmodel} and DrivingForward \citep{tian2025drivingforward}, we preserve their temporal modeling assumptions by treating the ego and collaborative observations as valid neighboring views within a short temporal window, again without modifying the core spatiotemporal architecture itself.

For scene-optimized baselines, the adaptation is performed at the scene level rather than at the network level. We export each cooperative sample into a unified scene representation that exposes synchronized multi-agent cameras and a shared ego-centric reference frame. V2X-Gaussians \citep{11097436} can then be trained on these exported scenes using its native cooperative formulation, while 3DGS \citep{kerbl20233d} and EmerNeRF \citep{yang2023emernerf} operate on the same scene representation as conventional multi-view reconstruction baselines. In this way, the comparison focuses on each method's actual modeling capacity under identical cooperative observations, rather than on extra engineering changes to the original network design.

\section{More Experimental Results}
\label{sec:appendix_more_results}

\subsection{Metric Definitions and Evaluation Modes}\label{sec:appendix_metrics}
We evaluate only the ego target views during benchmarking. When the input sequence contains four frames (two timestamps from ego and collab), the script evaluates only the ego frames. Let $\mathcal{I}_{pred}$ denote the rendered image, $\mathcal{I}_{gt}$ the corresponding GT target view, and $\mathbf{M}$ a binary evaluation mask over the image plane, where masked entries are set to $1$ and invalid entries are set to $0$.

\textbf{PSNR.} PSNR measures pixel-level reconstruction fidelity through the mean squared error. For full-image evaluation, the MSE is computed over all pixels. For masked settings such as \textit{Dynamic-only}, the binary mask is broadcast to all RGB channels, and the squared error is averaged strictly over valid masked entries. In Eq.~(\ref{eq:psnr}), $\odot$ denotes element-wise multiplication, and $\sum \mathbf{M}$ counts the total number of valid RGB entries after mask broadcasting. Higher PSNR indicates better reconstruction quality.
\begin{equation}
\text{MSE}(\mathcal{I}_{pred}, \mathcal{I}_{gt}; \mathbf{M}) =
\frac{\sum \big(\mathcal{I}_{pred} - \mathcal{I}_{gt}\big)^2 \odot \mathbf{M}}
{\sum \mathbf{M}},
\qquad
\text{PSNR} = -10 \log_{10}\big(\text{MSE}\big).
\label{eq:psnr}
\end{equation}

\textbf{SSIM.} SSIM evaluates local structural consistency instead of raw per-pixel error. Because an exact masked SSIM is difficult for patch-based metrics, the implementation adopts a region-focused approximation. Given a valid mask, the script first extracts the tight 2D bounding box of the masked region, expands it by 5 pixels on each side, and computes SSIM on the cropped patch when the crop is at least $11 \times 11$ pixels. In Eq.~(\ref{eq:ssim}), $(x_{\min}, x_{\max}, y_{\min}, y_{\max})$ denote the horizontal and vertical bounds of the padded bounding box. Higher SSIM is better.
\begin{equation}
\text{SSIM} =
\text{SSIM}\!\left(
\mathcal{I}_{pred}[y_{\min}\!:\!y_{\max}, x_{\min}\!:\!x_{\max}],
\mathcal{I}_{gt}[y_{\min}\!:\!y_{\max}, x_{\min}\!:\!x_{\max}]
\right).
\label{eq:ssim}
\end{equation}
This crop-based approximation is used to avoid the strong bias that zero padding would introduce in sparse masked regions.

\textbf{LPIPS.} LPIPS measures perceptual discrepancy in a deep feature space and is computed with an AlexNet-based LPIPS model in our script. Before evaluation, both images are linearly normalized from $[0,1]$ to $[-1,1]$, where $\tilde{\mathcal{I}}$ denotes the normalized image. For masked settings, the implementation extracts a 10-pixel padded bounding-box crop around the valid mask and computes LPIPS on that crop. In Eq.~(\ref{eq:lpips}), $d_{\text{Alex}}(\cdot,\cdot)$ denotes the AlexNet-based perceptual distance used by LPIPS. Lower LPIPS indicates better perceptual similarity.
\begin{equation}
\tilde{\mathcal{I}} = 2\mathcal{I} - 1,
\qquad
\text{LPIPS} = d_{\text{Alex}}\!\left(\tilde{\mathcal{I}}_{pred}, \tilde{\mathcal{I}}_{gt}\right).
\label{eq:lpips}
\end{equation}
Compared with direct masked zeroing, this cropped evaluation better reflects perceptual differences around dynamic objects.

\textbf{NIQE.} NIQE is used for \textit{Blind-Spot Completion}, where the true background is physically invisible and thus no reference image exists after removing the dynamic occluder. In this setting, the script evaluates the background-only rendering $\mathcal{I}_{bg}$, obtained by dropping dynamic Gaussians, inside the blind-spot region. The blind-spot mask $\mathbf{M}_{bs}$ is derived from the dynamic-object mask $\mathbf{M}_{dyn}$ as
\begin{equation}
\mathbf{M}_{bs} = \text{Clamp}\!\Big(\mathcal{D}_{\mathcal{K}_{bs}}(\mathbf{M}_{dyn}) - \mathbf{M}_{dyn},\, 0,\, 1\Big),
\label{eq:blindspot-mask}
\end{equation}
where $\mathbf{M}_{dyn}$ marks the image region occupied by dynamic objects, $\mathcal{D}_{\mathcal{K}_{bs}}(\cdot)$ denotes morphological dilation with a kernel $\mathcal{K}_{bs}$ of size $21 \times 21$, and $\text{Clamp}(\cdot,0,1)$ truncates all values to the valid interval $[0,1]$. Similar to LPIPS, NIQE is computed on a 10-pixel padded bounding-box crop around the valid blind-spot mask. In Eq.~(\ref{eq:niqe}), the cropped region is determined by $\mathbf{M}_{bs}$ and the coordinates $(x_{\min}, x_{\max}, y_{\min}, y_{\max})$ of its padded bounding box. Lower NIQE indicates that the completed region is more consistent with natural image statistics.
\begin{equation}
\text{NIQE} = \text{NIQE}\!\left(
\mathcal{I}_{bg}[y_{\min}\!:\!y_{\max}, x_{\min}\!:\!x_{\max}]
\right).
\label{eq:niqe}
\end{equation}
This design matches our blind-spot protocol, where only the plausibility of the completed hidden background can be assessed.

\textbf{Modes.} We report results under two temporal modes and three rendering settings. In \textit{Single-frame (SF)} mode, the target view at $t+1$ is extrapolated from observations at $t$. In \textit{Multi-frame (MF)} mode, the target view at $t$ is interpolated from observations at $t-1$ and $t+1$. Spatially, \textit{Full image} evaluates the complete rendered target view, \textit{Dynamic-only} focuses on the image region occupied by moving objects, and \textit{Blind-Spot Completion} focuses on the surrounding background region most strongly affected by dynamic occlusion.

\subsection{More Quantitative Results}

\noindent \textbf{Single-frame (SF) Mode NVS Results.} 
As discussed in the main text, we default to the multi-frame (MF) mode for our primary benchmark comparisons. For completeness, we additionally report the single-frame (SF) results for the baselines that can be fairly extended to SF evaluation. Specifically, the VGGT-based feed-forward methods, namely AnySplat, DGGT, and our \texttt{FRUC}, share the same predicted-pose extrapolation protocol: given the cooperative context at time $t_0$, the model first predicts the context camera parameters and scene representation, and the future ego target pose is then extrapolated from the predicted camera pose motion. DrivingForward is evaluated using its original native SF rendering branch under the same multi-agent input adaptation. The resulting comparison is shown in Table~\ref{tab:main_results_sf}.

\begin{table*}[t]
\centering
\caption{\textbf{Comparison of baselines in Single-Frame (SF) evaluation mode on V2X-Real.} The input context consists of both ego and collaborative views at $t$, and the task is to extrapolate the ego view at time $t+1$. \textbf{Best} in bold, and \underline{second-best} underlined.}
\label{tab:main_results_sf}
\footnotesize
\renewcommand{\arraystretch}{0.8}
\setlength{\tabcolsep}{4pt}
\resizebox{\textwidth}{!}{
\begin{tabular}{lccccccc}
\toprule
\multirow{2}{*}{Methods} & \multicolumn{3}{c}{Full image} & \multicolumn{3}{c}{Dynamic-only} & \multirow{2}{*}{\shortstack{Blind-Spot\\(NIQE$\downarrow$)}} \\
\cmidrule(lr){2-4} \cmidrule(lr){5-7}
& PSNR$\uparrow$ & SSIM$\uparrow$ & LPIPS$\downarrow$ & PSNR$\uparrow$ & SSIM$\uparrow$ & LPIPS$\downarrow$ & \\
\midrule
DrivingForward \citep{tian2025drivingforward} & 15.02 & \underline{0.682} & 0.289 & \underline{20.02} & \textbf{0.986} & \textbf{0.017} & - \\
AnySplat \citep{jiang2025anysplat} & \underline{19.04} & 0.603 & \underline{0.224} & 16.60 & \underline{0.979} & 0.031 & - \\
DGGT \citep{chen2025dggtfeedforward4dreconstruction} & 18.02 & 0.585 & 0.226 & 16.57 & 0.978 & 0.209 & \underline{6.021} \\
\midrule
\textbf{\texttt{FRUC}} (Ours) & \textbf{21.47} & \textbf{0.773} & \textbf{0.156} & \textbf{20.50} & \underline{0.979} & \underline{0.022} & \textbf{3.987} \\
\bottomrule
\vspace{-6mm}
\end{tabular}
}
\end{table*}

In SF evaluation mode, \texttt{FRUC} achieves the best overall performance, ranking first on all full-image metrics, Dynamic-only PSNR, and Blind-Spot NIQE. DrivingForward obtains the strongest Dynamic-only SSIM and LPIPS under its native SF branch, but its full-image fidelity remains clearly weaker, indicating that its future-view extrapolation is less stable at the scene level. By contrast, \texttt{FRUC} maintains consistently strong performance across both visible and occluded regions, which further supports its superior generalization and robustness in single-frame future-view prediction.

\textbf{Number of Input Collaborative Views.} 
To investigate the impact of collaborative information density, we compare DGGT and \texttt{FRUC} with different numbers of selected collaborative camera views on V2X-Real. In all settings, the ego input always contains two consecutive frames. $N_c=0$ denotes the ego-only setting, where the model receives only the two ego frames. $N_c=1$, which is also the default setting in the main paper, adds one associated collaborative camera across the same two timestamps, resulting in $2$ ego images plus $2$ collaborative images, i.e., $4$ inputs in total. $N_c=2$ further adds two associated collaborative cameras across the same two timestamps, resulting in $2$ ego images plus $4$ collaborative images, i.e., $6$ inputs in total. Table~\ref{tab:ablation_views} summarizes the results.

\begin{table*}[t]
\centering
\caption{\textbf{Effect of the Number of Input Collaborative Views on V2X-Real.} The ego input always contains two consecutive frames, while the number of collaborative inputs varies across settings.}
\label{tab:ablation_views}
\resizebox{\textwidth}{!}{
\begin{tabular}{lccccccccc}
\toprule
\multirow{2}{*}{\textbf{Method}} & \multicolumn{2}{c}{\textbf{Inputs}} & \multicolumn{3}{c}{\textbf{Full image}} & \multicolumn{3}{c}{\textbf{Dynamic-only}} & \multirow{2}{*}{\shortstack{\textbf{Blind-Spot}\\\textbf{(NIQE$\downarrow$)}}} \\
\cmidrule(lr){2-3} \cmidrule(lr){4-6} \cmidrule(lr){7-9}
& \textbf{Ego} & \textbf{Collab} & PSNR$\uparrow$ & SSIM$\uparrow$ & LPIPS$\downarrow$ & PSNR$\uparrow$ & SSIM$\uparrow$ & LPIPS$\downarrow$ & \\
\midrule
\multirow{3}{*}{\shortstack[c]{DGGT}} & 2 & 0 & 24.80 & 0.838 & 0.179 & 27.48 & 0.995 & 0.105 & 4.938 \\
 & 2 & 2 & 21.85 & 0.804 & 0.129 & 24.76 & 0.992 & 0.129 & 4.301 \\
 & 2 & 4 & 20.77 & 0.773 & 0.229 & 23.12 & 0.991 & 0.161 & 4.185 \\
\midrule
\multirow{3}{*}{\shortstack[c]{\texttt{FRUC}\\(Ours)}} & 2 & 0 & 27.02 & 0.893 & 0.006 & 27.93 & 0.995 & 0.005 & 4.985\\
 & 2 & 2 & 25.89 & 0.883 & 0.076 & 25.93 & 0.994 & 0.007 & 3.949 \\
 & 2 & 4 & 23.33 & 0.839 & 0.117 & 24.54 & 0.993 & 0.008 & 3.936 \\
\bottomrule
\end{tabular}
}
\vspace{-2mm}
\end{table*}

Compared with DGGT, \texttt{FRUC} better preserves the ego-only capability after full Stage II training, as reflected by its clearly stronger $N_c=0$ performance across both full-image and dynamic-only metrics. Moreover, \texttt{FRUC} remains more robust as additional collaborative views are introduced: although the task becomes increasingly challenging with larger cross-agent context, our method consistently maintains higher reconstruction quality and substantially better blind-spot completion, indicating a stronger capacity to exploit varying numbers of collaborative observations without severely corrupting the ego-centric representation.

\section{Theoretical Justification}\label{sec:appendix_cald}

In Section~\ref{sec:st_alignment}, we formulate cross-agent feature alignment as a constrained latent residual denoising process instantiated by the Cross-Agent Latent Residual Denoising (CALRD) module. In this section, we provide the formal theoretical justification for this design. We analyze the optimization bottleneck inherent in directly fine-tuning mixed latents under uncalibrated scenarios and mathematically demonstrate the convergence guarantees of our ego-conditioned residual formulation.

\subsection{Optimization Bottleneck in Direct Latent Fine-Tuning}

In the context of uncalibrated collaborative perception, a naive baseline directly aggregates multi-agent features to form a mixed latent $\tilde{\mathbf{F}}_{m}$ and subsequently fine-tunes the downstream Gaussian head $\Phi_{gs}$. Specifically, the mixed latent can be formulated as a corrupted observation:
\begin{equation}
\tilde{\mathbf{F}}_{m} = \mathbf{F}_{e} + \boldsymbol{\epsilon},
\end{equation}
where $\mathbf{F}_{e}$ denotes the structurally reliable ego-centric feature and $\boldsymbol{\epsilon}$ represents the non-stationary geometric noise induced by the highly irregular and unstable cross-agent spatio-temporal context. From a statistical machine learning perspective, direct fine-tuning constitutes an empirical risk minimization (ERM) problem under severely noisy covariates:
\begin{equation}
\min_{\theta} \mathcal{J}(\theta) = \mathbb{E}_{(\mathbf{F}_e, \boldsymbol{\epsilon}, Y)} \left[ \mathcal{L}\big(\Phi_{gs}(\mathbf{F}_e + \boldsymbol{\epsilon}), Y\big) \right],
\end{equation}
where $Y$ represents the target supervision signals (i.e., ground-truth images) and $\mathcal{L}$ is the rendering loss function.

To understand why optimizing this objective fails, we analyze the loss landscape via a second-order Taylor expansion around the clean ego-centric feature $\mathbf{F}_{e}$:
\begin{equation}
\mathcal{L}\big(\Phi_{gs}(\mathbf{F}_e + \boldsymbol{\epsilon}), Y\big) \approx \mathcal{L}\big(\Phi_{gs}(\mathbf{F}_e), Y\big) + \boldsymbol{\epsilon}^\top \nabla_{\mathbf{F}} \mathcal{L} + \frac{1}{2} \boldsymbol{\epsilon}^\top \mathbf{H}_{\mathbf{F}} \mathcal{L} \boldsymbol{\epsilon},
\end{equation}
where $\mathbf{H}_{\mathbf{F}}$ is the Hessian matrix of the loss with respect to the feature space. Because the uncalibrated cross-agent noise $\boldsymbol{\epsilon}$ exhibits high variance and lacks a consistent spatial structure, the expectation of the second-order noise penalty term $\mathbb{E} [\boldsymbol{\epsilon}^\top \mathbf{H}_{\mathbf{F}} \mathcal{L} \boldsymbol{\epsilon}]$ becomes excessively large. 

To minimize the overall empirical risk $\mathcal{J}(\theta)$, the optimizer is mathematically forced to reduce this severe noise penalty. According to Lipschitz continuity, doing so requires the network $\Phi_{gs}$ to learn a highly constrained mapping with a significantly suppressed local curvature, implicitly flattening the feature manifold to accommodate $\boldsymbol{\epsilon}$. Consequently, the model struggles to disentangle the cooperative signals from the misaligned noise. In its attempt to suppress the Hessian term, the optimizer inevitably compromises the optimization of the primary ideal risk $\mathcal{L}\big(\Phi_{gs}(\mathbf{F}_e), Y\big)$, distorting the ego vehicle's reliable local geometry originally encoded in $\mathbf{F}_e$. This phenomenon manifests as \textit{destructive semantic interference}, rendering the direct fine-tuning strategy mathematically unstable and practically infeasible.

\subsection{The Philosophy of FRUC}

To overcome the aforementioned optimization bottleneck, the FRUC architecture introduces a dedicated representation bottleneck, formalized as an ego-conditioned residual denoiser $\mathcal{R}_\phi$, situated between the pre-trained feature backbone and the downstream Gaussian head $\Phi_{gs}$. The core philosophy is to mathematically decouple the task of collaborative feature sanitization from the task of three-dimensional geometric decoding.

Instead of forcing the Gaussian head $\Phi_{gs}$ to directly map the out-of-distribution noisy latent $\tilde{\mathbf{F}}_{m}$ into a high-fidelity 3D Euclidean space, FRUC decomposes the original intractable empirical risk minimization into a constrained sequential optimization problem:
\begin{equation}
\min_{\phi, \theta} \mathbb{E} \left[ \mathcal{L}\Big(\Phi_{gs}\big(\underbrace{\tilde{\mathbf{F}}_{m} + \mathcal{R}_\phi(\tilde{\mathbf{F}}_{m}, \mathbf{F}_{e}, \mathbf{M}_{occ})}_{\hat{\mathbf{F}}_{m}}\big), Y\Big) \right].
\end{equation}
Here, the residual mapping $\mathcal{R}_\phi$ acts as a manifold projection operator. By explicitly conditioning this projection on the uncorrupted ego reference $\mathbf{F}_{e}$, the network establishes a deterministic structural anchor $\mathcal{M}_{ego} = \{ \mathbf{F} \mid \Vert \mathbf{F} - \mathbf{F}_e \Vert_2^2 \le \delta \}$. The denoised latent representation $\hat{\mathbf{F}}_{m}$ is forcibly pulled back onto this reliable ego-centric manifold, thereby guaranteeing that the Gaussian head $\Phi_{gs}$ receives topologically coherent tokens devoid of severe geometric ambiguities.

Furthermore, the explicit derivation of the occlusion uncertainty prior $\mathbf{M}_{occ} \in [0, 1]$ injects a critical spatial inductive bias that fundamentally resolves the inherent ambiguity of cross-agent completion. This spatial modulator elegantly partitions the optimization domain into two disjoint subspaces: the reliable visible region $\Omega_{vis} = \{ \mathbf{u} \mid \mathbf{M}_{occ}(\mathbf{u}) \to 0 \}$ and the dynamically occluded blind spot $\Omega_{occ} = \{ \mathbf{u} \mid \mathbf{M}_{occ}(\mathbf{u}) \to 1 \}$. This allows the optimization of the residual $\mathcal{R}_\phi$ to be mathematically decoupled:
\begin{equation}
\mathcal{R}_\phi(\mathbf{u}) \to 
\begin{cases} 
-\boldsymbol{\epsilon}(\mathbf{u}), & \mathbf{u} \in \Omega_{vis} \quad \text{(Driven by } \mathcal{L}_{den} \text{)}, \\
\Delta \mathbf{F}_{coop}(\mathbf{u}), & \mathbf{u} \in \Omega_{occ} \quad \text{(Driven by } \mathcal{L}_{coo} \text{)}.
\end{cases}
\end{equation}
In unoccluded regions $\Omega_{vis}$, the Ego-Reference Denoising Loss $\mathcal{L}_{den}$ mathematically constrains the residual mapping to approximate the negative noise $-\boldsymbol{\epsilon}$, effectively neutralizing the collaborative interference. Conversely, in dynamically occluded regions $\Omega_{occ}$, the Cooperative Completion Loss $\mathcal{L}_{coo}$ provides a targeted photometric gradient, driving the residual to synthesize missing cooperative structures $\Delta \mathbf{F}_{coop}$. 

Coupled with a strict zero-initialization strategy ($\mathcal{R}_\phi \to 0$ at $t=0$), this formulation guarantees a well-conditioned Jacobian matrix $\nabla_{\phi} \hat{\mathbf{F}}_{m} \approx \mathbf{I}$ during early training. Consequently, the FRUC architecture transforms the intractable direct mapping problem into a bounded, spatially modulated residual optimization, providing the Gaussian decoder with a mathematically stable and geometrically disambiguated hypothesis space to achieve superior rendering fidelity.

\subsection{The Effectiveness of Cross-Agent Latent Residual Denoising}

The rationale and effectiveness of formulating the Cross-Agent Latent Residual Denoising (CALRD) module specifically as an ego-conditioned residual denoising process, rather than relying on conventional cross-attention fusion, can be rigorously formalized using the Information Bottleneck (IB) principle. The fundamental objective of CALRD is to extract an optimal latent representation $\hat{\mathbf{F}}_{m}$ that maximizes predictive power for the target view $Y$ while minimizing the retention of irrelevant cross-agent geometric noise $\boldsymbol{\epsilon}$.

Traditional cross-attention mechanisms dynamically compute weighted combinations of multi-agent features. From an information-theoretic perspective, this operation expands the latent representation space, unintentionally maximizing the mutual information between the fused latent and the corrupted inputs. In highly uncalibrated scenarios, this expanded capacity becomes a liability, as the network is prone to memorizing spurious correlations.

In stark contrast, the CALRD paradigm explicitly acts as an information sink through a deterministic, single-step residual purification process. It is important to clarify that unlike generative Diffusion Models \citep{guo2026universalimagerestorationinternalized, ho2020denoisingdiffusionprobabilisticmodels}, which learn to reverse a multi-step stochastic Gaussian noise process to synthesize novel data, CALRD addresses a structured geometric misalignment problem. By mathematically treating the aggregated cooperative context $\tilde{\mathbf{F}}_{m} = \mathbf{F}_e + \boldsymbol{\epsilon}$ as a noisy observation of the true geometric features, the CALRD module aims to optimize the Information Bottleneck Lagrangian:
\begin{equation}
\mathcal{L}_{IB} = - I(\hat{\mathbf{F}}_{m}; Y) + \beta \cdot I(\hat{\mathbf{F}}_{m}; \tilde{\mathbf{F}}_{m}),
\end{equation}
where $I(\cdot; \cdot)$ denotes mutual information, and $\beta > 0$ is a trade-off multiplier. The first term encourages $\hat{\mathbf{F}}_{m}$ to retain task-relevant geometric structures for rendering $Y$, while the second term penalizes the complexity of the representation, forcing it to discard the non-stationary calibration noise $\boldsymbol{\epsilon}$ embedded within $\tilde{\mathbf{F}}_{m}$.

The structural design of the residual mapping $\mathcal{R}_\phi$ naturally enforces this compression. Because the mapping is strictly conditioned on the pure ego-reference $\mathbf{F}_{e}$, the mutual information can be decoupled:
\begin{equation}
I(\hat{\mathbf{F}}_{m}; \tilde{\mathbf{F}}_{m}) \approx I(\hat{\mathbf{F}}_{m}; \mathbf{F}_{e}) + I(\hat{\mathbf{F}}_{m}; \boldsymbol{\epsilon}).
\end{equation}
The Ego-Reference Denoising Loss $\mathcal{L}_{den}$ explicitly minimizes the second component $I(\hat{\mathbf{F}}_{m}; \boldsymbol{\epsilon})$ by driving $\mathcal{R}_\phi(\mathbf{u}) \to -\boldsymbol{\epsilon}(\mathbf{u})$ in visible regions $\Omega_{vis}$. Simultaneously, the Cooperative Completion Loss $\mathcal{L}_{coo}$ maximizes $I(\hat{\mathbf{F}}_{m}; Y)$ in the occluded blind spots $\Omega_{occ}$ by extracting necessary complementary structures $\Delta \mathbf{F}_{coop}$.

To ensure the optimization strictly evolves towards these constrained targets, CALRD follows the same dual-branch residual architecture described in Section~\ref{sec:st_alignment}. Specifically, the strong structural priors, namely the pure ego-reference $\mathbf{F}_{e}$ and the occlusion uncertainty mask $\mathbf{M}_{occ}$, are first encoded as
\begin{equation}
\mathbf{C}_{pri} = \mathcal{Z}_{pri} \Big( \Phi_{con}([\mathbf{M}_{occ}, \mathbf{F}_{e}]) \Big),
\end{equation}
and the residual correction is predicted as
\begin{equation}
\mathcal{R}_\phi(\tilde{\mathbf{F}}_{m}, \mathbf{F}_{e}, \mathbf{M}_{occ}) = \mathcal{Z}_{out} \Big( \Phi_{fea}(\tilde{\mathbf{F}}_{m}) + \mathbf{C}_{pri} \Big),
\end{equation}
so that the final denoised feature becomes $\hat{\mathbf{F}}_{m} = \tilde{\mathbf{F}}_{m} + \mathcal{R}_\phi(\tilde{\mathbf{F}}_{m}, \mathbf{F}_{e}, \mathbf{M}_{occ})$. Here, $\Phi_{con}$ is the lightweight condition encoder, $\Phi_{fea}$ is the feature encoder, and $\mathcal{Z}_{pri}$ and $\mathcal{Z}_{out}$ are zero-initialized $1 \times 1$ convolutions. Because the weights and biases of the zero-convolutions are strictly initialized to zero, the initial state of the residual correction satisfies $\mathcal{R}_\phi \approx 0$, which implies $\hat{\mathbf{F}}_{m} \approx \tilde{\mathbf{F}}_{m}$ at the beginning of training.

This architectural design is crucial for stable conditional guidance. During the early phases of gradient descent, the zero-initialization acts as a safe exploration mechanism. The network initially preserves the raw mixed latent $\tilde{\mathbf{F}}_{m}$ without adding uncontrolled perturbations. As training progresses, the gradients $\nabla_\phi \mathcal{L}_{IB}$ incrementally activate the zero-convolutions, allowing the network to cautiously learn the complex non-linear interactions between the noisy latent $\tilde{\mathbf{F}}_{m}$ and the structural priors. This mechanism mathematically guarantees that the injection of $\mathbf{M}_{occ}$ and $\mathbf{F}_{e}$ acts as a deterministic, progressively strengthening regularizer, reliably guiding the residual denoising process $\mathcal{R}_\phi$ along the steepest descent path towards the ideal information bottleneck optimum, explaining the empirical superiority and training stability of CALRD over naive end-to-end multi-agent fine-tuning strategies.


\end{document}